\documentclass[11pt]{article}
\usepackage[final]{acl}
\usepackage{times}
\usepackage{latexsym}
\usepackage[T1]{fontenc}
\usepackage[utf8]{inputenc}
\usepackage{microtype}
\usepackage{inconsolata}
\usepackage{graphicx}
\usepackage{amsmath}
\usepackage{amssymb}
\usepackage{longtable}
\usepackage{booktabs}
\usepackage{tabularx}
\usepackage{array}
\usepackage{enumitem}
\usepackage{fontawesome5}  
\usepackage{seqsplit}

\title{FPCO-Dialog: A Multi-Turn False-Premise Benchmark for Correction and Cooperation in Vision-Language Models}

\author{
    Jiayuan Ma$^1$, Yuqi Lu$^1$, Weiyang Guo$^1$, Chenrui Wang$^1$, Junyi Shu$^1$, \\ \textbf{Xuebo Liu}$^1$, \textbf{Min Zhang}$^1$, \textbf{Jing Li}$^1$\textsuperscript{\texorpdfstring{\faIcon[regular]{envelope}}{}} 
    \\$^{1}$Harbin Institute of Technology, Shenzhen, China  \\
    \texttt{anton.j.ma@outlook.com} \quad \texttt{jingli.phd@hotmail.com}  
}

\begin{document}
\maketitle
\begin{abstract}

Vision-language models (VLMs) are increasingly deployed in multi-turn settings where users may describe visual content with incorrect assumptions. Yet existing evaluations rarely isolate how models respond when the same visually grounded false premise persists across dialogue turns. We introduce FPCO-Dialog, a benchmark for evaluating correction and cooperation behavior in VLMs under repeated false premises. FPCO-Dialog contains 1,080 images and 10,800 question turns, stratified by visual complexity, object category, and false-premise class, and uses a 10-turn protocol in which a correct dialogue prefix is followed by repeated false-premise referring expressions. We evaluate 20 commercial and open-source VLMs with a model-agnostic protocol and CorrTP@K, a correction-rate metric over false-premise turns, scored by two independent detectors. FPCO-Dialog reveals substantial and persistent cross-model differences in aggregate correction tendency, model-specific turn-wise dynamics, and systematic variation across false-premise types under the benchmark's substitution distribution. The dataset, evaluation protocol, model outputs, detector labels, and code are available at \url{https://github.com/lab-klc/FPCO-Dialog}.
\let\thefootnote\relax\footnotetext{\faIcon[regular]{envelope}~Corresponding author.}

\end{abstract}

\begin{figure}[t]
    \centering
    \includegraphics[width=1\linewidth]{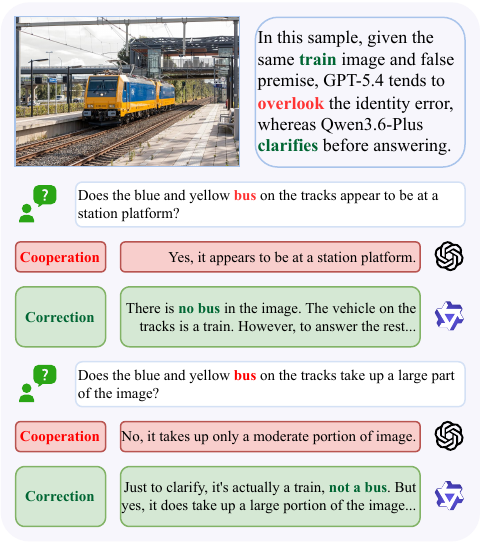}
    \caption{\textbf{Correction vs. Cooperation.} Under the same visually grounded false premise, VLMs may either cooperate with the user's main question or explicitly correct the premise before or while answering.}
    \label{fig:motivation}
\end{figure}

\section{Introduction}

Vision-language models (VLMs) are increasingly evaluated in settings that move beyond label prediction or short-answer visual question answering, including visually grounded interactive dialogue~\citep{visualdialog,visdiahalbench,multiverse}. Recent VLM benchmarks test broad multimodal competence, including integrated perception and reasoning, discipline-specific problem solving, hallucination robustness, and preference-aligned open-ended response quality~\citep{mmvet,mmmu,hallusionbench,wildvision}. Yet most evaluations either ask independent questions or use naturally collected dialogue histories, making it difficult to isolate how a model responds when a user repeatedly describes visible content using a premise that conflicts with the image~\citep{visualdialog,visdiahalbench,multiverse}. Figure~\ref{fig:motivation} illustrates this ambiguity: when a user refers to a train as a bus, a VLM may either cooperate with the user's main question or first correct the false premise. FPCO-Dialog is designed to measure this correction--cooperation behavior under a controlled repeated-false-premise protocol. A single-turn response may reveal whether a model notices one inconsistency, but repeated turns are needed to test whether correction or cooperation reflects a stable interaction strategy.

This setting is important because referring expressions do not merely identify an object; they also introduce presuppositions about the shared visual context, such as the identity, attribute, or location of a target~\citep{scorekeeping,commonground,presupposition}. In dialogue, such presuppositions may be accommodated for cooperation or explicitly repaired when incompatible with available evidence. Existing QA and VQA work has studied unanswerable questions, abstention, object hallucination, and false-premise questions, but these lines often frame problems as detecting unsupported answers rather than measuring the balance between correction and cooperation in visually grounded dialogue~\citep{know,vizwiz,reliable,pope,answering}. Recent visual-dialogue and hallucination benchmarks further show that dialogue history can affect VLM reliability, but they do not impose a controlled repeated same-false-premise protocol that separates premise type from image complexity and object category~\citep{visdiahalbench,hallusionbench,multiverse}.

We introduce FPCO-Dialog, a multi-turn false-premise benchmark for studying correction and cooperation behavior in VLMs. Each instance is built from an image selected from MS COCO and Open Images~\citep{coco,openimages} and paired with a ten-turn visual dialogue: an initial premise-correct prefix is followed by repeated turns using the same false-premise referring expression. The benchmark contains 1,080 images stratified by visual complexity, object category, and false-premise class, covering identity, attribute, and location errors. This design isolates how VLMs respond to persistent visually grounded false premises while keeping the dialogue protocol fixed and model-agnostic.

Our main contributions are as follows:
\begin{itemize}[noitemsep,nolistsep]
    \item We introduce FPCO-Dialog, a controlled and stratified multi-turn benchmark for repeated visually grounded false premises, covering 1,080 images across visual complexity, object category, and false-premise class.
    \item We propose a model-agnostic evaluation framework for measuring correction and cooperation behavior, including a fixed dialogue protocol, dual-detector scoring, a new indicator CorrTP@K, model outputs, detector labels, and evaluation code.
    \item We conduct a 20-model empirical study with quantitative and qualitative analyses which show substantial differences across model families and systematic variation across false-premise types.
\end{itemize}

\begin{figure*}[t]
    \centering
    \includegraphics[width=1\linewidth]{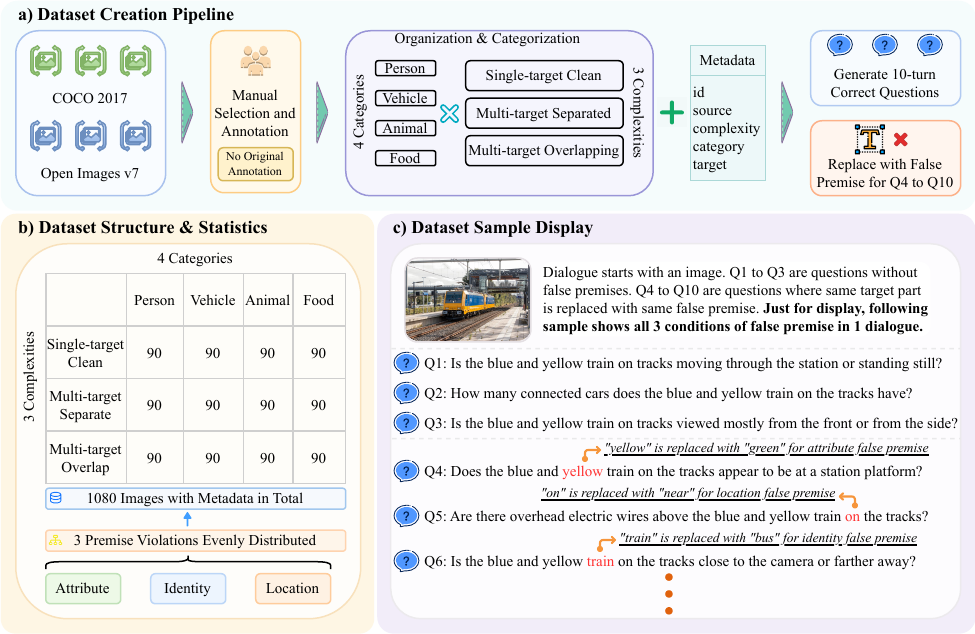}
    \caption{\textbf{Dataset Construction and Structure.} a) FPCO-Dialog is built from COCO and Open Images through manual selection, annotation, categorization, question generation, and false-premise rewriting. b) The dataset contains 1,080 images stratified by visual complexity, object category, and false-premise class. c) Each dialogue contains three premise-correct turns followed by repeated false-premise turns; the shown example illustrates all three false-premise classes for compact visualization.}
    \label{fig:dataset}
\end{figure*}

\section{Related Work}

\paragraph{Vision-language evaluation and visual dialogue.} Early visual question answering and visual dialogue benchmarks established image-conditioned question answering and dialogue grounding as core evaluation settings~\citep{vqa,visualdialog}. Recent VLM benchmarks extend this line to integrated perception and reasoning, expert-level multimodal problem solving, hallucination diagnosis, and preference-based open-ended evaluation~\citep{mmbench,mmvet,mmmu,hallusionbench,wildvision}. Evaluation toolkits further provide reproducible pipelines for comparing large multimodal models across benchmarks~\citep{vlmevalkit}. Recent multi-turn benchmarks study dialogue history, hallucination, and complex conversational goals in visually grounded interaction~\citep{visdiahalbench,multiverse}. FPCO-Dialog is complementary to these efforts: rather than maximizing task diversity, it fixes the dialogue protocol and systematically varies false-premise type, visual complexity, and object category to isolate correction behavior.

\paragraph{False premises, presupposition, and visually grounded inconsistency.} Our task is related to work on presupposition and common ground, where referring expressions can introduce assumptions that interlocutors may accommodate or repair depending on the conversational state~\citep{scorekeeping,commonground,presupposition}. In NLP and VQA, related phenomena have been studied through unanswerable questions, abstention-oriented answering, naturally unanswerable visual questions, and false-premise question answering~\citep{know,vizwiz,reliable,answering}. VLM hallucination benchmarks examine whether generated responses remain grounded in the image, especially when objects or visual facts are unsupported~\citep{pope,hallusionbench}. FPCO-Dialog differs in the direction and dynamics of inconsistency: the false content is supplied by the user, repeated across dialogue turns, and evaluated by whether the model corrects the premise or cooperates with the user's main question. FPCO-Dialog is also related to work on sycophancy, but we do not equate cooperation with sycophancy, since non-correction may reflect pragmatic accommodation or failure to detect the inconsistency rather than deference to the user~\citep{sharma2023understanding,hong2025sycophancy,pi2025sycophancy}.

\begin{figure*}[t]
    \centering
    \includegraphics[width=1\linewidth]{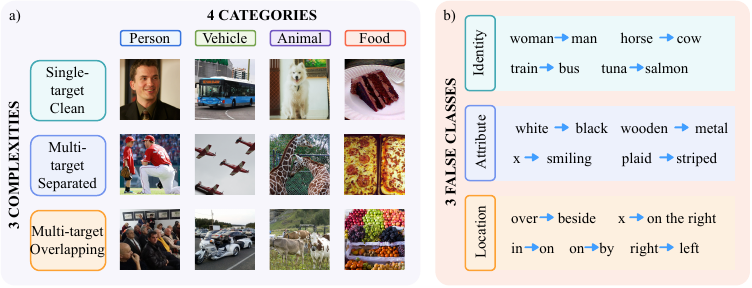}
    \caption{\textbf{Examples of Dataset Stratification and False-premise Modifications.} a) Representative FPCO-Dialog images across visual complexity and object category. Rows correspond to single-target clean, multi-target separated, and multi-target overlapping settings; columns correspond to person, vehicle, animal, and food categories. b) Examples of referring-expression modifications for the three false-premise classes: identity, attribute, and location.}
    \label{fig:sample}
\end{figure*}

\section{FPCO-Dialog Dataset}

FPCO-Dialog measures how VLMs respond to repeated visually grounded false premises. Each dialogue centers on a target entity in an image. Premise-correct turns refer to the target with a correct description, while false-premise turns use a modified referring expression. Figure~\ref{fig:dataset} summarizes construction, stratification, and a representative dialogue example.

\subsection{Task and Dialogue Protocol}

Each instance contains an image, a target entity, a correct target description, a modified false-premise target description, and a ten-turn question sequence. At each turn, a VLM receives the image, the current question, and the preceding dialogue context, then generates a free-form response.

The first three turns use the correct target description and form a premise-correct dialogue prefix. Turns 4--10 repeatedly use the same modified false-premise referring expression. This schedule isolates whether a model maintains, changes, or suppresses correction behavior when the same visual inconsistency persists across turns. On false-premise turns, a model may explicitly correct the premise, for example by stating that the referred object is absent or that the visual description is wrong. Alternatively, it may cooperate with the user's main question and answer without addressing the inconsistency. FPCO-Dialog measures this correction--cooperation distinction under a controlled repeated-false-premise protocol.

\subsection{Construction and Stratification}

We build FPCO-Dialog from images selected from MS COCO and Open Images~\citep{coco,openimages}. Images are manually selected and organized to support controlled comparisons across visual complexity, object category, and false-premise class. The authors define the benchmark design, stratification scheme, and construction constraints, while LLMs assist with question generation and false-premise rewriting. All generated instances are manually reviewed by the authors for visual grounding, target consistency, naturalness, and compliance with the single-error constraint. We do not rely on the original dataset annotations for benchmark labels or evaluation; target descriptions, question sequences, false-premise modifications, metadata, model outputs, and detector labels are generated or curated as part of FPCO-Dialog.

The dataset contains 1,080 images stratified along three axes. Visual complexity (Figure~\ref{fig:sample}a) has three levels: single-target clean, multi-target separated, and multi-target overlapping. Object category has four levels: person, vehicle, animal, and food. False-premise class has three levels: identity, attribute, and location. For each visual-complexity--category combination, we include 90 images, evenly divided across the three false-premise classes, yielding $3 \times 4 \times 3 \times 30 = 1{,}080$ images.

False-premise classes are defined by how the referring expression conflicts with the image. As shown in Figure~\ref{fig:sample}b, \textit{Identity} false premises replace the target identity with an incompatible alternative, such as referring to a train as a bus. \textit{Attribute} false premises modify a visible property of the target, such as color, material, or expression. \textit{Location} false premises modify the target's spatial description or relation, such as replacing ``right'' with ``left''. These examples are illustrative rather than fixed replacement mappings. For each instance, the substitution is generated independently under a constrained minimal-edit policy: we change one word when possible, otherwise one short local phrase, preserve the remaining description, and introduce only one error of the assigned false-premise class.

For each image, we first define one correct target description and generate ten premise-correct questions that contain this description. We then create one modified target description according to the image's assigned false-premise class and rewrite turns 4--10 by replacing the correct referring expression with the modified one. Identity substitutions use a clearly incompatible label, attribute substitutions change one visible property, and location substitutions change the spatial relation of the whole target. Thus, each image has exactly one target description, one modified target description, and one false-premise class. The example in Figure~\ref{fig:dataset} shows all three false-premise classes only for compact visualization; in the actual benchmark, each dialogue contains a single repeated false-premise class.

\subsection{Release and license}

We release the dataset, evaluation protocol, model outputs, detector labels, and code. Because the images are selected from MS COCO and Open Images, FPCO-Dialog is released as a mixed-license benchmark: source images remain subject to their original licenses, while author-generated components such as metadata, questions, false-premise modifications, evaluation scripts, and detector labels are released for research use.

\section{Experiment Setup}

\begin{figure*}[t]
    \centering
    \includegraphics[width=1\linewidth]{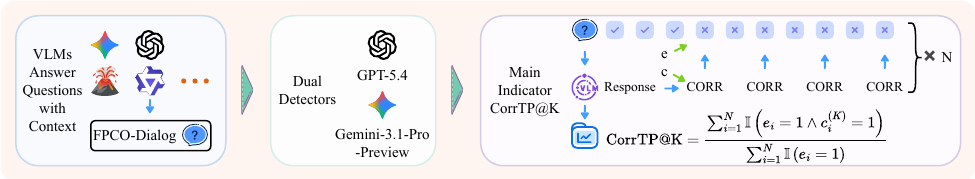}
    \caption{\textbf{Benchmark Procedure.} VLMs answer questions conditioned on image and dialogue context. Responses are scored by two detectors, and CorrTP@K measures correction behavior on false-premise turns.}
    \label{fig:benchmark}
\end{figure*}

\subsection{Evaluated Models}

\begin{figure}
    \centering
    \includegraphics[width=1\linewidth]{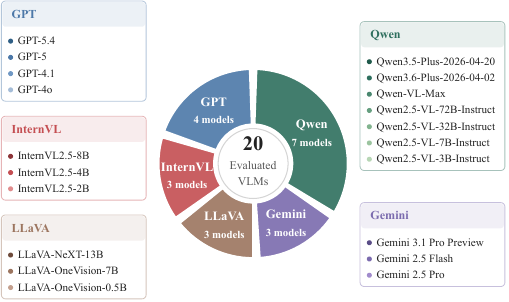}
    \caption{\textbf{Evaluated Models.} Twenty commercial/API-based and open-source VLMs are evaluated from five model families.}
    \label{fig:model}
\end{figure}

We evaluate 20 commercial/API-based and open-source VLMs from five model families: Gemini~\citep{gemini25flash,gemini25pro,gemini31pro}, GPT~\citep{gpt4o,gpt5,gpt41,gpt54}, Qwen~\citep{qwenvl,qwen25vl,qwen35plus,qwen36plus}, InternVL~\citep{internvl25}, and LLaVA~\citep{llava,llavaonevision}. As summarized in Figure~\ref{fig:model}, the model set covers frontier proprietary systems and open-source models across different scales. Each model produces one response for every question turn, yielding 10,800 responses per model and 216,000 model responses in total. We do not train or fine-tune any model; all experiments are inference-only evaluations. Model scales are reported where publicly available through model names or official documentation, while exact parameter counts and backend compute for proprietary API models are not publicly disclosed.

\subsection{Benchmarking Procedure}

Figure~\ref{fig:benchmark} summarizes the evaluation procedure. For each image, a model is evaluated on the ten-turn dialogue defined by FPCO-Dialog. At each turn, the model receives the image, the current user question, and the preceding dialogue context, then generates a free-form natural-language response. Responses are generated sequentially so that later turns include the earlier context in the dialogue history.

All models are evaluated with the same benchmark instances, turn order, and false-premise schedule. Models are not given the false-premise class, the correct target description, the modified target description, or detector labels. This prevents benchmark-side annotations from influencing response generation during inference. This setup helps isolate model behavior rather than differences in benchmark access. The protocol is therefore model-agnostic: any VLM that supports image-conditioned multi-turn responses can be evaluated by running the same dialogue and applying the same scoring pipeline. API-specific and local-inference scripts only adapt input formatting to each model interface; benchmark content and scoring are held fixed. For reproducibility, all model responses are generated with fixed inference settings within each model interface. We use deterministic decoding where supported, keep the maximum generation length fixed within each interface, and report detailed API, local inference, detector, and aggregation settings in Appendix~\ref{sec:prompts_used_in_benchmarking}. The same benchmark content and dialogue order are used for all models.

\subsection{Dual-detector Scoring}

Because model responses are open-ended, exact-match scoring is not suitable. Following LLM-as-judge and evaluator-model work~\citep{judging,prometheus}, as well as multimodal evaluation frameworks~\citep{vlmevalkit}, FPCO-Dialog uses detector-based scoring. Unlike general-purpose preference or quality judging, our detector task is restricted to a narrow behavioral criterion: whether the response corrects the false premise.

We use two independent detector models, GPT-5.4 and Gemini-3.1-Pro-Preview. Each detector receives, for each turn, the shared false-premise class, a flag indicating whether the turn contains a false premise, the premise-correct question, the corresponding false-premise question when applicable, and the model response. A response is labeled as a correction if it explicitly identifies the premise as inconsistent or clearly repairs the injected premise in its answer, such as stating that the referred object is absent, that the object is not the described entity, that a visible attribute or location is incorrect, or using the correct premise instead of the modified false premise. A response is not labeled as a correction if it simply answers the user's main question while accepting or ignoring the false premise. The main reported label is the arithmetic mean of the two binary detector labels.

\begin{figure*}[t]
    \centering
    \includegraphics[width=1\linewidth]{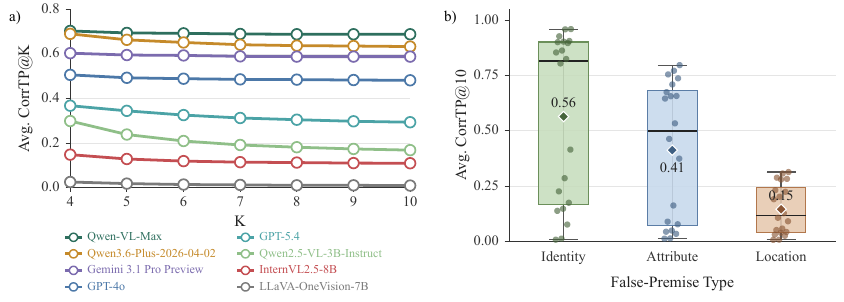}
    \caption{\textbf{Main Correction Behavior in FPCO-Dialog.} a) Representative models exhibit stable differences in average CorrTP@K across repeated false-premise turns. b) Correction behavior differs strongly by false-premise type, with identity errors corrected most often and location errors most often accommodated.}
    \label{fig:result1}
\end{figure*}

\begin{figure*}[t]
    \centering
    \includegraphics[width=1\linewidth]{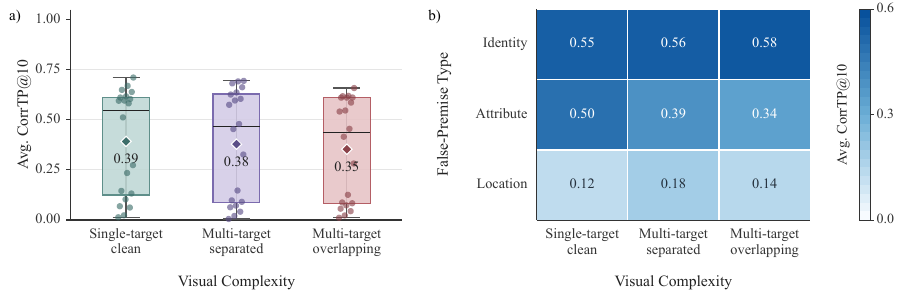}
    \caption{\textbf{Effects of Visual Complexity and Its Interaction with False-premise Type.} a) Average CorrTP@10 shows modest overall variation across visual complexity settings. b) Crossing visual complexity with false-premise type reveals clearer interaction patterns than the complexity-only view.}
    \label{fig:result2}
\end{figure*}

\subsection{Metrics}

We report CorrTP@K, a cumulative turn-indexed correction metric over false-premise turns up to dialogue position $K$. For each cutoff turn $K$, we collect all response items from turns up to $K$ and re-index them as $i=1,\ldots,N$. Let $e_i=1$ indicate that the selected response item contains a false premise, and let $c_i^{(K)}=1$ indicate that the response is labeled as a correction by the detector. CorrTP@K is defined as:

\[
\mathrm{CorrTP@K}
=
\frac{
\sum_{i=1}^{N}
\mathbb{I}\left(e_i = 1 \wedge c_i^{(K)} = 1\right)
}{
\sum_{i=1}^{N}
\mathbb{I}\left(e_i = 1\right)
}.
\]

CorrTP@K therefore measures the proportion of false-premise response items up to dialogue position $K$ for which the model explicitly corrects or clearly repairs the premise. Higher CorrTP@K indicates a stronger tendency toward explicit premise correction, but should not be interpreted as a general measure of response correctness, helpfulness, safety, or pragmatic appropriateness. In particular, cooperation denotes answering without explicitly repairing the injected premise and is not itself an error label; a corrected response may likewise still answer the user's main question.

As complementary metrics, we additionally report TurnCorr, a non-cumulative correction rate at each false-premise turn for analyzing turn-wise persistence or decay, and CorrFP@K, the correction rate on premise-correct turns for checking indiscriminate over-correction. Their definitions, formulas, and full model-level results are provided in Appendix~\ref{sec:extended_metrics}.

\section{Results}

We analyze CorrTP, i.e., correction on false-premise turns, using the arithmetic mean of the GPT-5.4 and Gemini-3.1-Pro-Preview detector labels. To assess the reliability of this detector-based scoring, we additionally validate the detector labels against expert human annotations on a stratified sample of 360 responses and observe strong human--detector agreement; the full validation protocol and results are reported in Appendix~\ref{sec:human_validation_of_llm_as_detectors}. To emphasize model separation and aggregate trends, the main text uses trajectory plots and aggregate summaries; full turn-indexed values for all models, detector-specific scores, and false-premise classes are provided in Appendix~\ref{sec:numbers_of_benchmark}.

\subsection{Correction Behavior across Models}

Figure~\ref{fig:result1}a shows cumulative average CorrTP@K trajectories for representative models from different families and scales. The curves are well separated, showing substantial differences in aggregate correction tendency across models. These differences remain relatively stable from $K=4$ to $K=10$, indicating persistent cross-model separation over repeated false-premise turns.

Because cumulative averaging can smooth turn-specific changes, we additionally examine non-cumulative TurnCorr@K in Appendix~\ref{sec:extended_metrics}. The turn-wise results show that some models maintain relatively stable correction rates, whereas others exhibit noticeable decay after the first false-premise turn. We therefore interpret CorrTP@K primarily as capturing stable cross-model differences in aggregate correction tendency, while TurnCorr@K reveals within-dialogue persistence or decay. As a specificity check, CorrFP is consistently low on the premise-correct prefix, with mean values of 0.0126, 0.0110, and 0.0101 at $K=1,2,3$ across models, suggesting that high CorrTP does not simply reflect indiscriminate correction (Appendix~\ref{sec:extended_metrics}).

\subsection{False-premise Type Matters}

Figure~\ref{fig:result1}b shows clear differences in correction behavior across false-premise types under the benchmark's current substitution distribution. Identity errors are corrected most often, with an average CorrTP@10 of approximately 0.56 across models. Attribute errors form an intermediate regime, with an average CorrTP@10 of approximately 0.41, while location errors are corrected least often, with an average CorrTP@10 of approximately 0.15.

We interpret this ordering as a descriptive pattern rather than a class-only causal effect, since correction rates may also depend on the particular lexical substitutions used. An exploratory lexical-sensitivity analysis with alternative substitutions shows that the separation of location from identity and attribute is relatively robust, whereas the identity--attribute difference is not robust to lexical variation (Appendix~\ref{sec:lexical_sensitivity}). In the current benchmark, location false premises are more frequently handled through cooperation, with models often answering the user's main question without explicitly correcting the spatial mismatch. Thus, class-level comparisons should be interpreted in conjunction with the benchmark's substitution distribution.

\subsection{Visual Complexity and Interaction Effects}

Figure~\ref{fig:result2}a shows that the main effect of visual complexity is modest. Average CorrTP@10 is approximately 0.39 for single-target clean images, 0.38 for multi-target separated images, and 0.35 for multi-target overlapping images.

Figure~\ref{fig:result2}b shows that identity errors remain relatively high across complexity settings, with average CorrTP@10 values of approximately 0.55, 0.56, and 0.58. Attribute errors decline from single-target clean images to more complex settings, from approximately 0.50 to 0.39 and 0.34. Location errors remain low in all settings, with values of approximately 0.12, 0.18, and 0.14.

These patterns show that visual complexity affects correction behavior differently across false-premise types. It has a clearer effect on attribute errors, where models must verify a local visual property of the target, while identity errors remain comparatively stable and location errors remain frequently handled through cooperation.

\begin{table*}[t]
\centering
\footnotesize
\setlength{\tabcolsep}{4pt}
\renewcommand{\arraystretch}{1.08}
\begin{tabularx}{\textwidth}{
@{}
>{\raggedright\arraybackslash}p{0.17\textwidth}
>{\raggedright\arraybackslash}p{0.23\textwidth}
>{\raggedright\arraybackslash}X
@{}}
\toprule
\textbf{Pattern} & \textbf{False-premise Setup} & \textbf{Representative Responses} \\
\midrule

\textbf{Cross-model contrast}
&
\textit{train} $\rightarrow$ \textit{bus}; asked about the ``bus on the tracks''
&
\textbf{GPT-5.4} (\textsc{Coop}): ``Yes, it appears to be at a station platform.'' \newline
\textbf{Qwen3.6-Plus} (\textsc{Corr}): ``There is no bus in the image. The vehicle on the tracks is a train...''
\\

&
\textit{pickup truck} $\rightarrow$ \textit{sedan}; asked facing direction
&
\textbf{GPT-5.4} (\textsc{Coop}): ``Toward the left.'' \newline
\textbf{Qwen3.6-Plus} (\textsc{Corr}): ``The vehicle is actually a pickup truck, not a sedan, but it is facing toward the left.''
\\

\midrule

\textbf{Correction while answering}
&
\textit{dog} $\rightarrow$ \textit{cat}; asked whether the ``cat'' has its mouth slightly open
&
\textbf{Qwen3.6-Plus} (\textsc{Corr}): ``The animal lying on the blue blanket is actually a \textbf{brindle dog}, not a cat. That said, yes...''
\\

&
\textit{red stripes} $\rightarrow$ \textit{blue stripes}; asked whether the airplane's wings are horizontal
&
\textbf{Qwen3.6-Plus} (\textsc{Corr}): ``The small propeller airplane actually has red stripes, not blue, but yes, its wings are stretched out horizontally.''
\\

\midrule

\textbf{Location cooperation}
&
\textit{on a black tray} $\rightarrow$ \textit{on a black tray near an oven}; asked about crust color
&
\textbf{Qwen3.6-Plus} (\textsc{Coop}): ``The crust of the large cheese pizza on the black tray is golden-brown...''
\\

&
\textit{gray road bicycle} $\rightarrow$ \textit{gray road bicycle on the right}; asked about drop handlebars
&
\textbf{Gemini 3.1 Pro Preview} (\textsc{Coop}): ``Yes, you can clearly see drop handlebars on the front of the gray road bicycle...''
\\

\midrule

\textbf{Attribute-specific model contrast}
&
\textit{green train} $\rightarrow$ \textit{red train}; asked whether it is on a curved section of rail
&
\textbf{GPT-5.4} (\textsc{Corr}): ``There is no red passenger train visible. The green train is on a curved section of track.'' \newline
\textbf{GPT-4o} (\textsc{Coop}): ``Yes, red passenger train is on curved section of rail.''
\\

\bottomrule
\end{tabularx}
\caption{\textbf{Representative case studies.}
The examples illustrate four recurring response patterns in FPCO-Dialog: different models may adopt different strategies under the same false premise, correction can still include an answer to the user's main question, location false premises are often handled through cooperation, and attribute errors can reveal model-dependent behavior. \textsc{Corr} denotes correction or clear repair of the false premise, and \textsc{Coop} denotes answering the main question without correcting it. Response excerpts are shortened for readability.}
\label{tab:case_studies}
\end{table*}

\subsection{Detector Consistency}

Because FPCO-Dialog scores open-ended responses, we verify that the two detectors produce consistent conclusions. The GPT-5.4 and Gemini-3.1-Pro-Preview detectors lead to similar model-family trends and the same overall ordering across false-premise classes. Some disagreements occur in borderline responses, where a model may mention the spatial mismatch but still answer the user's question. We therefore report the arithmetic mean of the two detector labels as the main score and provide agreement analyses in Appendix~\ref{sec:detectors_agreement_analyses}.

\section{Case Studies}

Table~\ref{tab:case_studies} illustrates representative response patterns behind the aggregate results. The first two rows show cross-model contrasts under the same identity false premise. In both examples, GPT-5.4 directly answers the user's main question under the modified referring expression, while Qwen3.6-Plus explicitly corrects the false premise before or while answering. These contrasts show that FPCO-Dialog captures differences in interaction strategy under matched visual and dialogue conditions, rather than only differences in task accuracy.

The third and fourth rows show that correction should not be interpreted as refusal. In these examples, the model repairs the false premise, identifying the animal as a dog rather than a cat or the airplane stripes as red rather than blue, but still answers the user's main question. This pattern is important for interpreting CorrTP@K: a higher correction rate reflects stronger premise-correction behavior, not a lower willingness to help or answer.

The fifth and sixth rows illustrate cooperation under location false premises. In the pizza and bicycle examples, the model answers the requested visual question without explicitly challenging the incorrect spatial phrase. This behavior is consistent with the quantitative finding that location false premises are frequently handled through cooperation, suggesting that models may treat some spatial mismatches as non-central to the main question.

The final row shows model-dependent behavior for an attribute false premise. Given the same red--green train mismatch, GPT-5.4 corrects the color premise, whereas GPT-4o answers under the modified description. This case complements the aggregate result that attribute errors occupy an intermediate regime: they are visually grounded and often correctable, but models differ substantially in whether they explicitly repair the premise.

\section{Conclusion}

We introduced FPCO-Dialog, a multi-turn benchmark for evaluating how VLMs respond to repeated visually grounded false premises. By fixing the dialogue protocol and systematically varying false-premise class, visual complexity, and object category, FPCO-Dialog enables controlled measurement and comparison of correction--cooperation behavior across models. Evaluating 20 commercial/API-based and open-source VLMs reveals substantial and persistent cross-model differences in aggregate correction tendency, together with model-specific turn-wise dynamics. Correction behavior also varies across false-premise types under the benchmark's current substitution distribution: identity errors are corrected most often on average, attribute errors form an intermediate regime, and location errors are frequently handled through cooperation. These findings suggest that VLM evaluation should consider not only whether models answer visual questions correctly, but also how they respond when user-supplied referring expressions conflict with visual evidence.

\section*{Limitations}

FPCO-Dialog isolates responses to repeated visually grounded false premises, but its fixed schedule does not cover all open-ended false-premise interactions. It covers three false-premise classes, four object categories, and English questions; broader linguistic, cultural, and domain coverage remains future work. Scoring uses two VLM-based detectors, so borderline cases may depend on detector interpretation. Evaluated behavior may change as systems and checkpoints are updated. Class-level differences may also depend on lexical, semantic, and contextual properties of the substitutions; our exploratory sensitivity analysis shows that the identity--attribute difference is not robust to lexical variation. CorrTP measures explicit premise repair rather than answer quality or pragmatic necessity, and does not distinguish harmless accommodation from failure to detect an inconsistency.

\section*{Ethical Considerations}

FPCO-Dialog is a research benchmark for VLM evaluation. It uses images from MS COCO and Open Images; source images remain under their original licenses. We release author-generated components---metadata, questions, false-premise modifications, evaluation protocol, model outputs, detector labels, and code---and document license requirements. Some images may contain people; we do not infer sensitive attributes, and all false premises are synthetic. We avoid adding names, unique identifiers, or offensive descriptions to author-generated content. All dataset selection, review, and curation are conducted by the authors. The human-validation annotations are also performed by two authors; no external annotators or crowdworkers are recruited.

\section*{Acknowledgements}
This work was supported in part by National Natural Science Foundation of China (62476070), Shenzhen Science and Technology Program \seqsplit{(JCYJ20241202123503005, \, GXWD20231128103232001, \,ZDSYS20230626091203008,\, KQTD20240729102154066)}, Department of Science and Technology of Guangdong (2024A1515011540). 

Large language models and vision-language models were used as auxiliary tools for dataset construction, evaluation, and limited writing assistance. Specifically, they assisted with question generation and false-premise rewriting, and GPT-5.4 and Gemini-3.1-Pro-Preview were used as detectors for automated scoring. These models were not used to define the research problem, formulate the benchmark design, choose the stratification axes, design the evaluation metrics, or draw conclusions from the results. All dataset design decisions, evaluation protocols, metric definitions, quantitative analyses, qualitative interpretations, and final findings were conducted and verified by the authors.

\bibliography{reference}
\appendix
\clearpage

\section{The Use of Large Language Models}

Large language models and vision-language models are used in this work as auxiliary tools for dataset construction, evaluation, and writing assistance. During dataset construction, they are used to help generate questions and rewrite expressions into false-premise variants. During evaluation, two vision-language-model-based detectors, GPT-5.4 and Gemini-3.1-Pro-Preview, are used to assign binary correction labels to open-ended model responses, as described in the experiment setup.

These models are not used to define the research problem, formulate the benchmark design, choose the stratification axes, design the evaluation metrics, or draw conclusions from the results. All dataset design decisions, evaluation protocols, metric definitions, quantitative analyses, qualitative interpretations, and final findings are conducted and verified by the authors.

\section{Detectors Agreement Analyses}
\label{sec:detectors_agreement_analyses}

The main CorrTP@K uses the arithmetic mean of two binary detector labels, from the GPT-5.4 and Gemini-3.1-Pro-Preview detectors. Because VLM responses are open-ended, correction behavior may appear in diverse forms, and relying on a single detector could introduce detector-specific bias. We therefore use two strong detectors from different model families and aggregate their binary labels.

Table~\ref{tab:detector_agreement_by_class} reports sample-level agreement on false-premise turns from $K=4$ to $K=10$. The two detectors show strong overall agreement, with an exact agreement rate of 0.973 and Cohen's $\kappa$ of 0.942. Agreement is highest for identity and attribute cases, while location cases are more ambiguous and show lower but still substantial agreement ($\kappa=0.809$). These results indicate that the dual-detector scoring procedure provides a stable basis for measuring correction behavior, while also motivating the human-validation analysis in Appendix~\ref{sec:human_validation_of_llm_as_detectors}.

\section{Human Validation of LLM as Detectors}
\label{sec:human_validation_of_llm_as_detectors}
\begin{table}[h]
\centering
\scriptsize
\setlength{\tabcolsep}{4pt}
\resizebox{\columnwidth}{!}{%
\begin{tabular}{lrrrr}
\toprule
Pair & Overall & Identity & Attribute & Location \\
\midrule
N & 360 & 120 & 120 & 120 \\
H1--H2 & 0.859 & 0.895 & 0.858 & 0.819 \\
H1--GPT & 0.835 & 0.864 & 0.856 & 0.827 \\
H1--Gemini & 0.832 & 0.897 & 0.821 & 0.805 \\
H2--GPT & 0.855 & 0.828 & 0.829 & 0.822 \\
H2--Gemini & 0.854 & 0.861 & 0.822 & 0.825 \\
GPT--Gemini & 0.960 & 0.966 & 0.983 & 0.856 \\
\bottomrule
\end{tabular}%
}
\caption{Pairwise Cohen's $\kappa$ on the human-validation subset. H1 and H2 denote the two domain-expert annotators from the author team. GPT and Gemini denote the GPT-5.4 and Gemini-3.1-Pro-Preview detectors, respectively.}
\label{tab:human_validation}
\end{table}

To validate the reliability of the LLM-based detectors, we conduct a stratified human evaluation on 360 model responses. For each of the 20 evaluated models, we randomly sample six responses from turns 4--10 for each of the three false-premise classes, yielding $20 \times 3 \times 6 = 360$ responses in total. Two domain-expert annotators from the author team independently label all sampled responses using the same binary correction criterion as the automated detectors. The annotators do not have access to the detector labels during annotation.

The annotators are instructed to label a response as a correction if it explicitly identifies the false premise as incorrect or clearly repairs the injected premise, and as a non-correction if it answers the main question while accepting or ignoring the false premise.

We measure pairwise agreement using Cohen's $\kappa$. As shown in Table~\ref{tab:human_validation}, the two human annotators achieve strong overall agreement ($\kappa=0.859$). Agreement between individual human annotators and the automated detectors is also high, ranging from $0.832$ to $0.855$ overall. Across the three false-premise classes, all human--detector agreement scores remain above $0.80$. The two automated detectors also show high agreement on this human-validation subset, with an overall Cohen's $\kappa$ of $0.960$.

These results indicate that both GPT-5.4 and Gemini-3.1-Pro-Preview closely track expert human judgments under the correction criterion used in FPCO-Dialog. Human annotation is used here to validate the automated scoring procedure rather than as a required component of benchmark evaluation, allowing the benchmark to retain an automated and scalable evaluation pipeline.

\section{Judge-Bias and Self-Scoring Analysis}
\label{sec:judge_bias_analysis}

Because GPT-5.4 and Gemini-3.1-Pro-Preview serve as both detectors and evaluated models, we examine potential self-scoring bias. The detector prompt does not reveal the identity of the evaluated model. We further perform a leave-one-judge-out check by scoring each detector model using only the other detector. For GPT-5.4, the score changes from 0.2939 under dual-detector scoring to 0.2886 using Gemini alone, with its rank unchanged at 12. For Gemini-3.1-Pro-Preview, the score changes from 0.5880 to 0.5906 using GPT alone, with its rank unchanged at 7. The overall 20-model ranking is also unchanged under this check, indicating that the reported comparisons are not materially affected by detector self-scoring.

\section{Numbers in Benchmark}
\label{sec:numbers_of_benchmark}

This section reports the numerical results underlying the turn-level and category-level analyses in the main text. While the main figures emphasize trajectory patterns and aggregate comparisons, Table~\ref{tab:full_corrtp_numbers} provides exact cumulative CorrTP@K values for each evaluated model on false-premise turns. We report detector-specific scores from GPT-5.4 and Gemini-3.1-Pro-Preview, together with their arithmetic mean, for the overall benchmark and each false-premise class. Since the first three turns form the premise-correct prefix, only turns $K=4$ to $K=10$ are included in this table. FPCO-Dialog is used as an evaluation benchmark, and we do not define train, development, or test splits.

\section{Extended Metrics}
\label{sec:extended_metrics}

In addition to the cumulative CorrTP@K metric used in the main analysis, we report TurnCorr to characterize correction behavior at individual false-premise turns (detailed numbers in Table~\ref{tab:full_turncorr_numbers}). Unlike CorrTP@K, which aggregates all false-premise responses up to turn $K$, TurnCorr uses only the responses at the current dialogue turn $K$. This makes it possible to distinguish persistent correction from turn-specific decay or other changes that may be smoothed by cumulative averaging. For a false-premise turn $K$, we collect all response items at that turn and re-index them as $i=1,\ldots,N$. Let $e_i=1$ indicate that the selected response item contains an injected false premise, and let $c_i^{(K)}=1$ indicate that the model response is labeled as a correction or clear premise repair by the detector. We define TurnCorr@K as:

\begin{equation}
\mathrm{TurnCorr@K}
\frac{
\sum_{i=1}^{N}
\mathbb{I}\left(e_i=1 \wedge c_i^{(K)}=1\right)
}{
\sum_{i=1}^{N}
\mathbb{I}\left(e_i=1\right)
}.
\end{equation}

We also report CorrFP@K as a specificity check for indiscriminate correction (detailed numbers in Table~\ref{tab:full_corrfp_numbers}). CorrFP@K measures how often a model produces a correction on premise-correct turns, where no false premise has been injected. A low CorrFP@K therefore indicates that a model's correction behavior is targeted toward actual false premises rather than reflecting a general tendency to challenge user descriptions. For a cutoff turn $K$, we collect all response items from turns up to $K$ and re-index them as $i=1,\ldots,N$. Let $e_i=1$ indicate that the selected response item contains an injected false premise, and let $e_i=0$ indicate that it is premise-correct. Let $c_i^{(K)}=1$ indicate that the model response is labeled as a correction or clear premise repair by the detector. We define CorrFP@K as:

\begin{equation}
\mathrm{CorrFP@K}
\frac{
\sum_{i=1}^{N}
\mathbb{I}\left(e_i=0 \wedge c_i^{(K)}=1\right)
}{
\sum_{i=1}^{N}
\mathbb{I}\left(e_i=0\right)
}.
\end{equation}

In the current FPCO-Dialog protocol, TurnCorr@K is reported for the false-premise turns $K=4,\ldots,10$, while CorrFP@K is evaluated on the premise-correct prefix $K=1,2,3$. Together, these metrics complement CorrTP@K by respectively exposing turn-level correction dynamics and checking for false-positive premise correction.

\section{Lexical Sensitivity Analysis}
\label{sec:lexical_sensitivity}

To examine whether the class-level differences depend on the particular lexical substitutions used in FPCO-Dialog, we conduct an exploratory matched sensitivity study on 30 unique images. For each false-premise class, we construct two alternative substitution sets (A and B) while holding the image, target, premise-correct dialogue prefix, and Q4 question intent fixed. We evaluate three models from different model families. Candidate variants are additionally checked by VLMs from three independent families without access to model responses, detector labels, or the expected class ordering.

As shown in Table~\ref{tab:lexical_sensitivity}, both substitution sets produce the same descriptive ordering, with location premises receiving lower correction rates than identity and attribute premises. However, across 1,000 random selections between the A/B alternatives, the full Identity $>$ Attribute $>$ Location ordering is retained in only 73.1\% of runs. The separation of location from identity and attribute is comparatively robust, whereas the identity--attribute difference is sensitive to lexical variation. The strict ordering also holds for only two of the three evaluated models. We therefore treat the class ordering in the main benchmark as a descriptive pattern under the current substitution distribution rather than a class-only causal effect. This small-scale analysis is exploratory and does not fully control semantic or perceptual salience.

\begin{table}[h]
\centering
\scriptsize
\setlength{\tabcolsep}{4pt}
\resizebox{\columnwidth}{!}{%
\begin{tabular}{lrrrr}
\toprule
Set & Identity & Attribute & Location & Ordering \\
\midrule
A      & 0.722 & 0.700 & 0.633 & I $>$ A $>$ L \\
B      & 0.783 & 0.761 & 0.539 & I $>$ A $>$ L \\
Pooled & 0.753 & 0.731 & 0.586 & I $>$ A $>$ L \\
\bottomrule
\end{tabular}%
}
\caption{Exploratory lexical-sensitivity results under two alternative substitution sets. Values denote average correction rates across the three evaluated models.}
\label{tab:lexical_sensitivity}
\end{table}

\section{Protocol Robustness}
\label{sec:protocol_robustness}

We do not claim that the fixed three-turn premise-correct prefix, seven false-premise turns, or the current dataset size are uniquely optimal. We therefore examine whether the aggregate findings depend strongly on these protocol choices. First, truncating the false-premise sequence shows that the descriptive Identity $>$ Attribute $>$ Location ordering is already present at $K=4$ and remains unchanged at every cutoff through $K=10$. At the same time, the non-cumulative TurnCorr analysis reveals meaningful within-dialogue dynamics: for 10 of the 20 evaluated models, TurnCorr decreases by more than five percentage points from Q4 to Q10, with the largest decrease reaching 15.93 points. Thus, shorter protocols can recover coarse aggregate patterns, while repeated turns provide additional information about persistence and decay.

We also conduct an image-level stratified bootstrap over the 36 visual-complexity--category--false-premise cells. Using 10, 15, 20, 25, or 30 sampled images per cell, with 1,000 repetitions for each setting, the Identity $>$ Attribute $>$ Location ordering is retained in all repetitions. A corresponding stratified sampling analysis without replacement yields the same result. These analyses indicate that the aggregate class-level pattern is stable to substantially smaller stratified samples under the current substitution distribution; they should not be interpreted as establishing an optimal sample size or as controlling lexical variation, which is examined separately in Appendix~\ref{sec:lexical_sensitivity}.

\section{Prompts and Inference Settings Used in Benchmarking}
\label{sec:prompts_used_in_benchmarking}

This section summarizes the main prompts used in the FPCO-Dialog pipeline. We include prompts for question generation, false-premise rewriting, benchmark inference, and detector-based scoring. The templates below are lightly normalized for readability: implementation-specific API formatting, image encoding, and credential-related details are omitted. Variables are shown in braces. This appendix also reports the API interfaces, local model-loading procedures, decoding settings, detector-output formats, and aggregation settings used to produce the reported results.

\paragraph{Question generation prompt.}
The question-generation prompt is used to construct a target-centric premise-correct dialogue for each image. The prompt asks the model to first identify a single visible target and then generate ten questions that all contain the same target description.

\begin{quote}
\small
You are writing a controlled multi-turn question set for one image.

Input image file: \texttt{\{image\_name\}}

Metadata: \texttt{\{metadata\_text\}}

Requirements:
\begin{enumerate}
    \item First identify one visible target and write one \texttt{target\_description} for it.
    \item The \texttt{target\_description} should be objective, correct, short, and natural.
    \item Keep the \texttt{target\_description} compact enough to fit smoothly inside a question. Use about two to four useful descriptive details.
    \item Then write exactly \texttt{\{QUESTION\_COUNT\}} correct questions about that same target.
    \item All questions must stay with the same target. Do not switch to another person, object, animal, or background element.
    \item Every question must contain the full \texttt{target\_description} exactly as written, word for word.
    \item Do not paraphrase, shorten, or replace the \texttt{target\_description} with pronouns or vague references.
    \item Use the \texttt{target\_description} exactly once in each question.
    \item Let the rest of each question vary naturally around that fixed target description.
    \item Keep the questions grounded in what can be reasonably asked about the target in the image, such as visible state, action, nearby relation, or role in the scene.
    \item Keep the wording natural and conversational.
    \item Use the target indicated by the metadata. For the category number in metadata, 1 means person, 2 means vehicle, 3 means animal, and 4 means food.
\end{enumerate}

Return JSON with a \texttt{target\_description} field and a \texttt{questions} array, where each question has an \texttt{id} and \texttt{content}.
\end{quote}

\paragraph{False-premise rewriting prompt.}
The false-premise rewriting prompt is used to create the repeated false-premise turns. For each image, the false-premise class is fixed to one of identity, attribute, or location. The prompt asks the model to minimally modify the original target description and then replace that description consistently in turns $K=4$ to $K=10$.

\begin{quote}
\small
You are creating controlled false-premise questions for one image.

Metadata: \texttt{\{metadata\_text\}}

Original \texttt{target\_description}: \texttt{\{target\_description\}}

False-premise questions to modify: \texttt{\{items\_text\}}

Return only the final JSON.

The false class for this image is fixed: \texttt{\{false\_class\}}

Requirements:
\begin{enumerate}
    \item First choose one \texttt{modified\_target\_description} based on the original \texttt{target\_description}.
    \item Change only one word if possible. If one word is not natural, change only one short local phrase.
    \item Keep the rest of the target description unchanged.
    \item Do not expand the description.
    \item Do not create multiple errors.
    \item The result should still sound like a natural referring expression.
    \item The changed part must match the fixed false class.
    \item For identity edits, use a clearly incompatible identity substitution. Avoid age-graded, near-synonym, vague, or broader/narrower label changes such as \textit{woman} $\rightarrow$ \textit{girl}, \textit{man} $\rightarrow$ \textit{boy}, or \textit{person} $\rightarrow$ \textit{woman}. Prefer a firmer identity change such as \textit{woman} $\leftrightarrow$ \textit{man}, \textit{dog} $\leftrightarrow$ \textit{cat}, or \textit{car} $\leftrightarrow$ \textit{truck} when it fits naturally.
    \item For location edits, make the change a natural position or relative-position change of the whole target. The \texttt{modified\_target\_description} must remain a complete referring expression that can be placed directly into a question. Attach the location relation to the target and a scene place or scene object, not to clothing, attributes, body parts, or local descriptive phrases.
    \item For each question, replace the original \texttt{target\_description} with the same \texttt{modified\_target\_description}.
    \item Keep each question's intent and wording otherwise as unchanged as possible.
    \item If an identity edit creates an obvious pronoun mismatch, adjust only those pronouns to match the new identity or use a natural neutral phrasing.
    \item Do not rewrite questions for style.
\end{enumerate}

False class definitions:
\begin{itemize}
    \item \textbf{identity}: change who or what the target is with a clearly incompatible identity, not a soft label shift.
    \item \textbf{attribute}: change one property of the target.
    \item \textbf{location}: change where the whole target is or its relative position in the scene.
\end{itemize}

This \texttt{modified\_target\_description} will be reused across multiple turns for the same image, so keep it stable and concise.

Return JSON with a \texttt{modified\_target\_description} field and a \texttt{questions} array, where each item contains an \texttt{id} and the rewritten \texttt{modified} question.
\end{quote}

\paragraph{Benchmark inference prompt.}
All evaluated models use the same semantic inference prompt. Differences across GPT, Gemini, Qwen, InternVL, and LLaVA scripts only adapt the image encoding, chat template, or API message format required by each model interface. At inference time, benchmarked models receive the image, the current question, and the prior dialogue history.

\begin{quote}
\small
System prompt:

\textit{You are a helpful assistant. Answer the current user question based on the image and the prior conversation history.}

Dialogue template:

First user turn:
\[
\texttt{[image input]} \quad \texttt{\{question\_1\}}
\]

Assistant:
\[
\texttt{\{response\_1\}}
\]

User:
\[
\texttt{\{question\_2\}}
\]

Assistant:
\[
\texttt{\{response\_2\}}
\]

\noindent
$\cdots$

Current user turn:
\[
\texttt{\{current\_question\}}
\]
\end{quote}

The image is included with the first user turn, and previous turns are included as alternating user and assistant messages. The benchmarked models are not given the false-premise class, the correct target description, the modified target description, detector labels, or scoring metadata in the inference prompt.

\paragraph{Detector scoring prompt.}
The two detector models use the same semantic scoring prompt. The detector receives the shared false-premise class, the premise-correct question, the modified false-premise question when applicable, and the model response. The output is a binary label indicating whether the response corrects or clearly repairs the premise. The detectors do not receive the image and are not asked to independently verify visual facts; they only judge whether the response corrects or clearly repairs the known injected premise.

\begin{quote}
\small
You are labeling model responses for the FPCO-Dialog benchmark.

Use the shared \texttt{false\_class}, \texttt{false\_premise}, \texttt{content}, \texttt{modified}, and \texttt{response} fields.

Decide whether the response should be labeled \texttt{detect = true} or \texttt{detect = false}.

Decision rule:

\begin{enumerate}
    \item If \texttt{false\_premise = true}:
    \begin{itemize}
        \item \texttt{detect = true} if the response explicitly says the premise is wrong.
        \item \texttt{detect = true} if the response says the described object, person, attribute, or location is not present or not correct.
        \item \texttt{detect = true} if the response directly corrects the user, such as saying it is not X but Y.
        \item \texttt{detect = true} if the response clearly uses the correct premise from \texttt{content} instead of following the false premise in \texttt{modified}.
        \item \texttt{detect = false} only when the response goes along with the false premise in \texttt{modified} and does not show correction.
    \end{itemize}

    \item If \texttt{false\_premise = false}:
    \begin{itemize}
        \item \texttt{detect = false} for a normal answer to the correct question.
        \item \texttt{detect = true} only if the response unnecessarily questions, corrects, or rejects a question that was actually correct.
    \end{itemize}
\end{enumerate}

How to use \texttt{false\_class}:
\begin{itemize}
    \item \textbf{identity}: the modified question changes who or what the target is.
    \item \textbf{attribute}: the modified question changes one property of the target.
    \item \textbf{location}: the modified question changes where the target is or its relative position.
\end{itemize}

Return labels for every provided id. Return JSON only.
\end{quote}

The GPT-5.4 detector uses a strict JSON schema requiring an integer \texttt{id} and a boolean \texttt{detect} field. The Gemini-3.1-Pro-Preview detector uses the same semantic prompt with JSON output parsing. The main CorrTP@K results use the arithmetic mean of the two detector labels, and detector agreement analyses are reported in Appendix~\ref{sec:detectors_agreement_analyses}.

\paragraph{Inference and implementation settings.}
All experiments are inference-only evaluations; we do not train, fine-tune, or perform hyperparameter search for any evaluated model. For benchmark inference, we use deterministic decoding where supported. Gemini API inference uses temperature $0.0$, top-$p=1.0$, and \texttt{maxOutputTokens=256} by default, with Pro-family Gemini models using at least 1024 output tokens. GPT API inference uses \texttt{max\_output\_tokens=256} by default, increased to at least 512 for GPT-5-family models; temperature and top-$p$ are only sent when a positive temperature is requested. Qwen API inference uses \texttt{max\_tokens=256}; temperature and top-$p$ are likewise only sent when a positive temperature is requested. Local InternVL, LLaVA, and Qwen inference use PyTorch and Hugging Face Transformers with \texttt{max\_new\_tokens=256} and default temperature $0.0$; under the default setting, \texttt{do\_sample=False}. No benchmark inference script sets top-$k$, a fixed random seed, quantization, or an explicit batch size.

Detector scoring uses two detector models with the same semantic labeling prompt. The GPT-5.4 detector uses the OpenAI Responses API with strict JSON-schema output requiring an integer \texttt{id} and a boolean \texttt{detect} label. The Gemini-3.1-Pro-Preview detector uses temperature $0.0$, top-$p=1.0$, \texttt{maxOutputTokens=8192}, and JSON MIME output. The main CorrTP@K, TurnCorr@K, and CorrFP@K statistics are deterministic aggregations of saved model responses and do not use random sampling or multiple stochastic runs; the separate stratified bootstrap used for protocol robustness is described in Appendix~\ref{sec:protocol_robustness}.

\clearpage
\onecolumn
\begin{table*}[h]
\centering
\scriptsize
\setlength{\tabcolsep}{4pt}
\resizebox{\textwidth}{!}{%
\begin{tabular}{lrrrrrrrrr}
\toprule
Scope & N & GPT Corr. & Gemini Corr. & Agree & Both Corr. & Both Non-Corr. & GPT Only & Gemini Only & $\kappa$ \\
\midrule
Overall & 151200 & 0.379 & 0.370 & 0.973 & 0.361 & 0.612 & 0.018 & 0.009 & 0.942 \\
Identity & 50400 & 0.563 & 0.566 & 0.986 & 0.557 & 0.428 & 0.006 & 0.009 & 0.971 \\
Attribute & 50400 & 0.415 & 0.411 & 0.981 & 0.404 & 0.577 & 0.012 & 0.007 & 0.960 \\
Location & 50400 & 0.159 & 0.133 & 0.952 & 0.122 & 0.830 & 0.037 & 0.011 & 0.809 \\
\bottomrule
\end{tabular}%
}
\caption{Detector agreement by false-premise class on false-premise turns. GPT Corr. and Gemini Corr. denote the correction rates assigned by the GPT-5.4 and Gemini-3.1-Pro-Preview detectors, respectively. Agree denotes exact agreement between the two binary detector labels. Both Corr. and Both Non-Corr. indicate cases where both detectors assign correction and non-correction labels, respectively. GPT Only and Gemini Only denote one-sided correction labels.}
\label{tab:detector_agreement_by_class}
\end{table*}
\begingroup
\small
\setlength{\tabcolsep}{2.2pt}
\renewcommand{\arraystretch}{1.04}

\begin{longtable}{@{}lrrrrrrrrrrrr@{}}
\toprule
K & O-GPT & O-Gem & O-Avg & Id-GPT & Id-Gem & Id-Avg & Attr-GPT & Attr-Gem & Attr-Avg & Loc-GPT & Loc-Gem & Loc-Avg \\
\midrule
\endfirsthead

\toprule
K & O-GPT & O-Gem & O-Avg & Id-GPT & Id-Gem & Id-Avg & Attr-GPT & Attr-Gem & Attr-Avg & Loc-GPT & Loc-Gem & Loc-Avg \\
\midrule
\endhead

\multicolumn{13}{c}{\textbf{Gemini 2.5 Flash}} \\
4 & 0.650 & 0.641 & 0.645 & 0.928 & 0.944 & 0.936 & 0.756 & 0.753 & 0.754 & 0.267 & 0.225 & 0.246 \\
5 & 0.661 & 0.645 & 0.653 & 0.946 & 0.956 & 0.951 & 0.768 & 0.761 & 0.764 & 0.269 & 0.218 & 0.243 \\
6 & 0.660 & 0.646 & 0.653 & 0.952 & 0.959 & 0.956 & 0.769 & 0.764 & 0.766 & 0.260 & 0.215 & 0.237 \\
7 & 0.658 & 0.644 & 0.651 & 0.953 & 0.958 & 0.956 & 0.769 & 0.763 & 0.766 & 0.253 & 0.210 & 0.231 \\
8 & 0.660 & 0.646 & 0.653 & 0.957 & 0.959 & 0.958 & 0.772 & 0.766 & 0.769 & 0.252 & 0.212 & 0.232 \\
9 & 0.660 & 0.646 & 0.653 & 0.957 & 0.959 & 0.958 & 0.773 & 0.767 & 0.770 & 0.251 & 0.212 & 0.231 \\
10 & 0.661 & 0.647 & 0.654 & 0.958 & 0.960 & 0.959 & 0.775 & 0.768 & 0.772 & 0.250 & 0.213 & 0.231 \\

\midrule
\multicolumn{13}{c}{\textbf{Gemini 2.5 Pro}} \\
4 & 0.704 & 0.688 & 0.696 & 0.947 & 0.947 & 0.947 & 0.811 & 0.800 & 0.806 & 0.353 & 0.317 & 0.335 \\
5 & 0.681 & 0.669 & 0.675 & 0.939 & 0.942 & 0.940 & 0.776 & 0.767 & 0.772 & 0.326 & 0.300 & 0.313 \\
6 & 0.673 & 0.662 & 0.667 & 0.936 & 0.938 & 0.937 & 0.766 & 0.759 & 0.762 & 0.316 & 0.288 & 0.302 \\
7 & 0.662 & 0.652 & 0.657 & 0.931 & 0.935 & 0.933 & 0.751 & 0.745 & 0.748 & 0.303 & 0.276 & 0.289 \\
8 & 0.657 & 0.650 & 0.653 & 0.928 & 0.932 & 0.930 & 0.744 & 0.742 & 0.743 & 0.299 & 0.275 & 0.287 \\
9 & 0.655 & 0.647 & 0.651 & 0.927 & 0.931 & 0.929 & 0.741 & 0.740 & 0.740 & 0.297 & 0.272 & 0.284 \\
10 & 0.653 & 0.646 & 0.649 & 0.925 & 0.929 & 0.927 & 0.738 & 0.736 & 0.737 & 0.294 & 0.273 & 0.283 \\

\midrule
\multicolumn{13}{c}{\textbf{Gemini 3.1 Pro Preview}} \\
4 & 0.606 & 0.601 & 0.603 & 0.808 & 0.822 & 0.815 & 0.647 & 0.658 & 0.653 & 0.361 & 0.322 & 0.342 \\
5 & 0.598 & 0.592 & 0.595 & 0.807 & 0.817 & 0.812 & 0.643 & 0.653 & 0.648 & 0.343 & 0.306 & 0.325 \\
6 & 0.596 & 0.590 & 0.593 & 0.802 & 0.811 & 0.806 & 0.644 & 0.652 & 0.648 & 0.341 & 0.306 & 0.324 \\
7 & 0.591 & 0.586 & 0.589 & 0.801 & 0.808 & 0.804 & 0.641 & 0.649 & 0.645 & 0.332 & 0.301 & 0.317 \\
8 & 0.591 & 0.585 & 0.588 & 0.800 & 0.807 & 0.804 & 0.642 & 0.649 & 0.645 & 0.329 & 0.299 & 0.314 \\
9 & 0.591 & 0.585 & 0.588 & 0.800 & 0.808 & 0.804 & 0.643 & 0.649 & 0.646 & 0.329 & 0.299 & 0.314 \\
10 & 0.591 & 0.585 & 0.588 & 0.801 & 0.808 & 0.804 & 0.643 & 0.648 & 0.645 & 0.328 & 0.300 & 0.314 \\

\midrule
\multicolumn{13}{c}{\textbf{GPT-4.1}} \\
4 & 0.619 & 0.611 & 0.615 & 0.917 & 0.917 & 0.917 & 0.689 & 0.678 & 0.683 & 0.253 & 0.239 & 0.246 \\
5 & 0.610 & 0.599 & 0.605 & 0.914 & 0.912 & 0.913 & 0.683 & 0.674 & 0.679 & 0.233 & 0.211 & 0.222 \\
6 & 0.604 & 0.594 & 0.599 & 0.910 & 0.911 & 0.911 & 0.676 & 0.669 & 0.673 & 0.226 & 0.203 & 0.215 \\
7 & 0.599 & 0.591 & 0.595 & 0.907 & 0.909 & 0.908 & 0.670 & 0.662 & 0.666 & 0.219 & 0.200 & 0.210 \\
8 & 0.595 & 0.586 & 0.591 & 0.904 & 0.907 & 0.905 & 0.666 & 0.657 & 0.661 & 0.216 & 0.194 & 0.205 \\
9 & 0.593 & 0.584 & 0.589 & 0.903 & 0.906 & 0.905 & 0.663 & 0.654 & 0.659 & 0.213 & 0.193 & 0.203 \\
10 & 0.592 & 0.583 & 0.587 & 0.901 & 0.906 & 0.903 & 0.662 & 0.651 & 0.657 & 0.213 & 0.191 & 0.202 \\

\midrule
\multicolumn{13}{c}{\textbf{GPT-4o}} \\
4 & 0.523 & 0.490 & 0.506 & 0.878 & 0.872 & 0.875 & 0.489 & 0.467 & 0.478 & 0.203 & 0.131 & 0.167 \\
5 & 0.512 & 0.475 & 0.493 & 0.864 & 0.863 & 0.863 & 0.478 & 0.461 & 0.470 & 0.193 & 0.103 & 0.148 \\
6 & 0.508 & 0.471 & 0.489 & 0.860 & 0.858 & 0.859 & 0.474 & 0.459 & 0.467 & 0.190 & 0.094 & 0.142 \\
7 & 0.505 & 0.467 & 0.486 & 0.858 & 0.856 & 0.857 & 0.473 & 0.458 & 0.466 & 0.184 & 0.087 & 0.136 \\
8 & 0.504 & 0.465 & 0.485 & 0.857 & 0.855 & 0.856 & 0.472 & 0.457 & 0.465 & 0.183 & 0.083 & 0.133 \\
9 & 0.503 & 0.463 & 0.483 & 0.857 & 0.855 & 0.856 & 0.471 & 0.456 & 0.464 & 0.181 & 0.080 & 0.131 \\
10 & 0.502 & 0.462 & 0.482 & 0.855 & 0.854 & 0.855 & 0.471 & 0.456 & 0.464 & 0.179 & 0.077 & 0.128 \\

\midrule
\multicolumn{13}{c}{\textbf{GPT-5}} \\
4 & 0.072 & 0.075 & 0.073 & 0.125 & 0.133 & 0.129 & 0.069 & 0.083 & 0.076 & 0.022 & 0.008 & 0.015 \\
5 & 0.071 & 0.076 & 0.073 & 0.121 & 0.131 & 0.126 & 0.076 & 0.090 & 0.083 & 0.015 & 0.007 & 0.011 \\
6 & 0.072 & 0.076 & 0.074 & 0.126 & 0.131 & 0.129 & 0.076 & 0.090 & 0.083 & 0.015 & 0.007 & 0.011 \\
7 & 0.074 & 0.079 & 0.076 & 0.131 & 0.136 & 0.134 & 0.079 & 0.093 & 0.086 & 0.013 & 0.007 & 0.010 \\
8 & 0.075 & 0.079 & 0.077 & 0.132 & 0.137 & 0.135 & 0.081 & 0.094 & 0.087 & 0.013 & 0.006 & 0.009 \\
9 & 0.076 & 0.080 & 0.078 & 0.134 & 0.139 & 0.137 & 0.083 & 0.097 & 0.090 & 0.012 & 0.005 & 0.009 \\
10 & 0.077 & 0.081 & 0.079 & 0.137 & 0.141 & 0.139 & 0.083 & 0.097 & 0.090 & 0.011 & 0.005 & 0.008 \\

\midrule
\multicolumn{13}{c}{\textbf{GPT-5.4}} \\
4 & 0.376 & 0.360 & 0.368 & 0.575 & 0.542 & 0.558 & 0.433 & 0.425 & 0.429 & 0.119 & 0.114 & 0.116 \\
5 & 0.351 & 0.339 & 0.345 & 0.533 & 0.508 & 0.520 & 0.411 & 0.404 & 0.407 & 0.108 & 0.104 & 0.106 \\
6 & 0.332 & 0.319 & 0.326 & 0.491 & 0.469 & 0.480 & 0.397 & 0.391 & 0.394 & 0.109 & 0.098 & 0.104 \\
7 & 0.319 & 0.307 & 0.313 & 0.462 & 0.442 & 0.452 & 0.390 & 0.384 & 0.387 & 0.106 & 0.094 & 0.100 \\
8 & 0.311 & 0.299 & 0.305 & 0.446 & 0.428 & 0.437 & 0.384 & 0.379 & 0.382 & 0.102 & 0.089 & 0.096 \\
9 & 0.304 & 0.293 & 0.298 & 0.431 & 0.416 & 0.423 & 0.381 & 0.375 & 0.378 & 0.099 & 0.087 & 0.093 \\
10 & 0.299 & 0.289 & 0.294 & 0.422 & 0.409 & 0.415 & 0.377 & 0.372 & 0.374 & 0.098 & 0.085 & 0.091 \\

\midrule
\multicolumn{13}{c}{\textbf{InternVL2.5-2B}} \\
4 & 0.139 & 0.131 & 0.135 & 0.178 & 0.175 & 0.176 & 0.094 & 0.092 & 0.093 & 0.144 & 0.128 & 0.136 \\
5 & 0.099 & 0.093 & 0.096 & 0.138 & 0.131 & 0.135 & 0.058 & 0.058 & 0.058 & 0.100 & 0.089 & 0.095 \\
6 & 0.079 & 0.074 & 0.076 & 0.115 & 0.109 & 0.112 & 0.045 & 0.046 & 0.045 & 0.077 & 0.067 & 0.072 \\
7 & 0.069 & 0.064 & 0.067 & 0.099 & 0.094 & 0.097 & 0.044 & 0.043 & 0.043 & 0.063 & 0.053 & 0.058 \\
8 & 0.062 & 0.057 & 0.059 & 0.091 & 0.086 & 0.088 & 0.039 & 0.038 & 0.038 & 0.057 & 0.047 & 0.052 \\
9 & 0.058 & 0.052 & 0.055 & 0.085 & 0.080 & 0.083 & 0.038 & 0.036 & 0.037 & 0.050 & 0.041 & 0.045 \\
10 & 0.054 & 0.049 & 0.052 & 0.079 & 0.075 & 0.077 & 0.035 & 0.034 & 0.035 & 0.048 & 0.038 & 0.043 \\

\midrule
\multicolumn{13}{c}{\textbf{InternVL2.5-4B}} \\
4 & 0.145 & 0.134 & 0.140 & 0.217 & 0.211 & 0.214 & 0.106 & 0.103 & 0.104 & 0.114 & 0.089 & 0.102 \\
5 & 0.110 & 0.102 & 0.106 & 0.185 & 0.179 & 0.182 & 0.074 & 0.069 & 0.072 & 0.072 & 0.058 & 0.065 \\
6 & 0.096 & 0.090 & 0.093 & 0.169 & 0.167 & 0.168 & 0.061 & 0.058 & 0.059 & 0.056 & 0.044 & 0.050 \\
7 & 0.087 & 0.082 & 0.084 & 0.160 & 0.158 & 0.159 & 0.054 & 0.052 & 0.053 & 0.046 & 0.036 & 0.041 \\
8 & 0.081 & 0.077 & 0.079 & 0.153 & 0.152 & 0.152 & 0.049 & 0.047 & 0.048 & 0.040 & 0.032 & 0.036 \\
9 & 0.078 & 0.075 & 0.076 & 0.150 & 0.150 & 0.150 & 0.047 & 0.045 & 0.046 & 0.037 & 0.030 & 0.034 \\
10 & 0.076 & 0.073 & 0.074 & 0.148 & 0.148 & 0.148 & 0.045 & 0.043 & 0.044 & 0.035 & 0.028 & 0.032 \\

\midrule
\multicolumn{13}{c}{\textbf{InternVL2.5-8B}} \\
4 & 0.152 & 0.147 & 0.149 & 0.283 & 0.281 & 0.282 & 0.103 & 0.100 & 0.102 & 0.069 & 0.061 & 0.065 \\
5 & 0.131 & 0.126 & 0.129 & 0.250 & 0.250 & 0.250 & 0.089 & 0.086 & 0.087 & 0.056 & 0.043 & 0.050 \\
6 & 0.123 & 0.118 & 0.120 & 0.241 & 0.240 & 0.240 & 0.082 & 0.080 & 0.081 & 0.044 & 0.034 & 0.039 \\
7 & 0.117 & 0.113 & 0.115 & 0.234 & 0.233 & 0.234 & 0.080 & 0.077 & 0.079 & 0.037 & 0.028 & 0.033 \\
8 & 0.114 & 0.111 & 0.113 & 0.231 & 0.231 & 0.231 & 0.078 & 0.076 & 0.077 & 0.034 & 0.026 & 0.030 \\
9 & 0.113 & 0.109 & 0.111 & 0.229 & 0.228 & 0.229 & 0.077 & 0.075 & 0.076 & 0.031 & 0.024 & 0.028 \\
10 & 0.112 & 0.108 & 0.110 & 0.227 & 0.227 & 0.227 & 0.079 & 0.077 & 0.078 & 0.030 & 0.022 & 0.026 \\

\midrule
\multicolumn{13}{c}{\textbf{LLaVA-NeXT-13B}} \\
4 & 0.165 & 0.148 & 0.157 & 0.292 & 0.289 & 0.290 & 0.094 & 0.083 & 0.088 & 0.108 & 0.072 & 0.090 \\
5 & 0.140 & 0.120 & 0.130 & 0.247 & 0.242 & 0.244 & 0.074 & 0.065 & 0.070 & 0.099 & 0.053 & 0.076 \\
6 & 0.124 & 0.106 & 0.115 & 0.217 & 0.214 & 0.215 & 0.065 & 0.056 & 0.060 & 0.091 & 0.048 & 0.070 \\
7 & 0.113 & 0.098 & 0.106 & 0.200 & 0.199 & 0.200 & 0.061 & 0.053 & 0.057 & 0.077 & 0.042 & 0.059 \\
8 & 0.107 & 0.093 & 0.100 & 0.190 & 0.188 & 0.189 & 0.059 & 0.050 & 0.054 & 0.072 & 0.039 & 0.055 \\
9 & 0.102 & 0.089 & 0.096 & 0.182 & 0.180 & 0.181 & 0.056 & 0.049 & 0.053 & 0.069 & 0.037 & 0.053 \\
10 & 0.099 & 0.085 & 0.092 & 0.177 & 0.174 & 0.175 & 0.054 & 0.046 & 0.050 & 0.068 & 0.036 & 0.052 \\

\midrule
\multicolumn{13}{c}{\textbf{LLaVA-OneVision-0.5B}} \\
4 & 0.060 & 0.051 & 0.055 & 0.047 & 0.053 & 0.050 & 0.025 & 0.039 & 0.032 & 0.108 & 0.061 & 0.084 \\
5 & 0.044 & 0.032 & 0.038 & 0.029 & 0.028 & 0.029 & 0.019 & 0.026 & 0.022 & 0.085 & 0.043 & 0.064 \\
6 & 0.039 & 0.027 & 0.033 & 0.022 & 0.020 & 0.021 & 0.016 & 0.021 & 0.019 & 0.078 & 0.039 & 0.058 \\
7 & 0.035 & 0.022 & 0.029 & 0.019 & 0.017 & 0.018 & 0.015 & 0.019 & 0.017 & 0.071 & 0.031 & 0.051 \\
8 & 0.032 & 0.019 & 0.026 & 0.016 & 0.014 & 0.015 & 0.015 & 0.018 & 0.017 & 0.066 & 0.027 & 0.046 \\
9 & 0.030 & 0.018 & 0.024 & 0.014 & 0.013 & 0.013 & 0.015 & 0.017 & 0.016 & 0.061 & 0.024 & 0.042 \\
10 & 0.028 & 0.016 & 0.022 & 0.013 & 0.012 & 0.013 & 0.013 & 0.015 & 0.014 & 0.057 & 0.022 & 0.040 \\

\midrule
\multicolumn{13}{c}{\textbf{LLaVA-OneVision-7B}} \\
4 & 0.021 & 0.031 & 0.026 & 0.019 & 0.042 & 0.030 & 0.025 & 0.039 & 0.032 & 0.019 & 0.014 & 0.017 \\
5 & 0.016 & 0.021 & 0.019 & 0.017 & 0.025 & 0.021 & 0.018 & 0.026 & 0.022 & 0.014 & 0.011 & 0.013 \\
6 & 0.014 & 0.017 & 0.015 & 0.012 & 0.019 & 0.015 & 0.017 & 0.023 & 0.020 & 0.012 & 0.008 & 0.010 \\
7 & 0.012 & 0.014 & 0.013 & 0.010 & 0.015 & 0.013 & 0.014 & 0.019 & 0.017 & 0.011 & 0.007 & 0.009 \\
8 & 0.011 & 0.012 & 0.011 & 0.008 & 0.012 & 0.010 & 0.013 & 0.017 & 0.015 & 0.013 & 0.006 & 0.009 \\
9 & 0.011 & 0.011 & 0.011 & 0.007 & 0.011 & 0.009 & 0.012 & 0.016 & 0.014 & 0.013 & 0.006 & 0.009 \\
10 & 0.010 & 0.010 & 0.010 & 0.007 & 0.010 & 0.009 & 0.011 & 0.015 & 0.013 & 0.012 & 0.005 & 0.009 \\

\midrule
\multicolumn{13}{c}{\textbf{Qwen2.5-VL-32B-Instruct}} \\
4 & 0.629 & 0.620 & 0.625 & 0.894 & 0.900 & 0.897 & 0.717 & 0.700 & 0.708 & 0.275 & 0.261 & 0.268 \\
5 & 0.601 & 0.597 & 0.599 & 0.871 & 0.876 & 0.873 & 0.689 & 0.683 & 0.686 & 0.243 & 0.232 & 0.237 \\
6 & 0.590 & 0.590 & 0.590 & 0.869 & 0.874 & 0.871 & 0.677 & 0.678 & 0.677 & 0.226 & 0.219 & 0.223 \\
7 & 0.582 & 0.584 & 0.583 & 0.863 & 0.870 & 0.867 & 0.668 & 0.672 & 0.670 & 0.214 & 0.212 & 0.213 \\
8 & 0.576 & 0.579 & 0.577 & 0.862 & 0.870 & 0.866 & 0.664 & 0.666 & 0.665 & 0.202 & 0.203 & 0.203 \\
9 & 0.572 & 0.575 & 0.573 & 0.860 & 0.866 & 0.863 & 0.659 & 0.662 & 0.661 & 0.197 & 0.196 & 0.197 \\
10 & 0.570 & 0.572 & 0.571 & 0.860 & 0.867 & 0.863 & 0.658 & 0.657 & 0.657 & 0.191 & 0.191 & 0.191 \\

\midrule
\multicolumn{13}{c}{\textbf{Qwen2.5-VL-3B-Instruct}} \\
4 & 0.316 & 0.283 & 0.299 & 0.425 & 0.417 & 0.421 & 0.328 & 0.303 & 0.316 & 0.194 & 0.131 & 0.163 \\
5 & 0.250 & 0.229 & 0.239 & 0.362 & 0.362 & 0.362 & 0.244 & 0.228 & 0.236 & 0.143 & 0.096 & 0.119 \\
6 & 0.218 & 0.201 & 0.210 & 0.331 & 0.332 & 0.332 & 0.208 & 0.195 & 0.202 & 0.113 & 0.076 & 0.095 \\
7 & 0.199 & 0.184 & 0.192 & 0.313 & 0.312 & 0.312 & 0.190 & 0.178 & 0.184 & 0.094 & 0.062 & 0.078 \\
8 & 0.189 & 0.175 & 0.182 & 0.302 & 0.301 & 0.301 & 0.182 & 0.171 & 0.176 & 0.083 & 0.054 & 0.069 \\
9 & 0.180 & 0.168 & 0.174 & 0.293 & 0.293 & 0.293 & 0.172 & 0.162 & 0.167 & 0.076 & 0.048 & 0.062 \\
10 & 0.175 & 0.163 & 0.169 & 0.287 & 0.286 & 0.286 & 0.167 & 0.159 & 0.163 & 0.072 & 0.044 & 0.058 \\

\midrule
\multicolumn{13}{c}{\textbf{Qwen2.5-VL-72B-Instruct}} \\
4 & 0.647 & 0.622 & 0.635 & 0.900 & 0.900 & 0.900 & 0.747 & 0.733 & 0.740 & 0.294 & 0.233 & 0.264 \\
5 & 0.635 & 0.614 & 0.625 & 0.900 & 0.900 & 0.900 & 0.742 & 0.731 & 0.736 & 0.263 & 0.212 & 0.237 \\
6 & 0.637 & 0.615 & 0.626 & 0.900 & 0.900 & 0.900 & 0.754 & 0.742 & 0.748 & 0.256 & 0.205 & 0.230 \\
7 & 0.634 & 0.614 & 0.624 & 0.901 & 0.900 & 0.901 & 0.754 & 0.742 & 0.748 & 0.249 & 0.201 & 0.225 \\
8 & 0.635 & 0.615 & 0.625 & 0.903 & 0.903 & 0.903 & 0.756 & 0.744 & 0.750 & 0.247 & 0.199 & 0.223 \\
9 & 0.636 & 0.616 & 0.626 & 0.904 & 0.904 & 0.904 & 0.758 & 0.745 & 0.752 & 0.246 & 0.198 & 0.222 \\
10 & 0.639 & 0.618 & 0.629 & 0.906 & 0.906 & 0.906 & 0.762 & 0.749 & 0.756 & 0.250 & 0.200 & 0.225 \\

\midrule
\multicolumn{13}{c}{\textbf{Qwen2.5-VL-7B-Instruct}} \\
4 & 0.610 & 0.563 & 0.587 & 0.919 & 0.906 & 0.913 & 0.625 & 0.594 & 0.609 & 0.286 & 0.189 & 0.237 \\
5 & 0.560 & 0.527 & 0.544 & 0.878 & 0.869 & 0.873 & 0.594 & 0.572 & 0.583 & 0.208 & 0.140 & 0.174 \\
6 & 0.533 & 0.506 & 0.520 & 0.852 & 0.847 & 0.849 & 0.570 & 0.550 & 0.560 & 0.178 & 0.122 & 0.150 \\
7 & 0.516 & 0.495 & 0.506 & 0.838 & 0.837 & 0.837 & 0.557 & 0.540 & 0.548 & 0.154 & 0.108 & 0.131 \\
8 & 0.506 & 0.488 & 0.497 & 0.832 & 0.832 & 0.832 & 0.548 & 0.532 & 0.540 & 0.139 & 0.099 & 0.119 \\
9 & 0.500 & 0.483 & 0.491 & 0.825 & 0.829 & 0.827 & 0.544 & 0.528 & 0.536 & 0.133 & 0.093 & 0.113 \\
10 & 0.497 & 0.480 & 0.488 & 0.823 & 0.826 & 0.825 & 0.542 & 0.526 & 0.534 & 0.126 & 0.088 & 0.107 \\

\midrule
\multicolumn{13}{c}{\textbf{Qwen3.5-Plus-2026-04-20}} \\
4 & 0.680 & 0.661 & 0.671 & 0.933 & 0.939 & 0.936 & 0.758 & 0.744 & 0.751 & 0.347 & 0.300 & 0.324 \\
5 & 0.651 & 0.632 & 0.641 & 0.918 & 0.925 & 0.921 & 0.714 & 0.697 & 0.706 & 0.322 & 0.274 & 0.298 \\
6 & 0.639 & 0.622 & 0.631 & 0.902 & 0.914 & 0.908 & 0.696 & 0.682 & 0.689 & 0.319 & 0.269 & 0.294 \\
7 & 0.631 & 0.614 & 0.623 & 0.897 & 0.906 & 0.901 & 0.687 & 0.673 & 0.680 & 0.309 & 0.262 & 0.285 \\
8 & 0.628 & 0.611 & 0.619 & 0.893 & 0.906 & 0.899 & 0.687 & 0.672 & 0.679 & 0.303 & 0.256 & 0.279 \\
9 & 0.625 & 0.610 & 0.617 & 0.890 & 0.903 & 0.897 & 0.684 & 0.669 & 0.677 & 0.302 & 0.257 & 0.279 \\
10 & 0.624 & 0.611 & 0.617 & 0.888 & 0.905 & 0.897 & 0.683 & 0.667 & 0.675 & 0.300 & 0.260 & 0.280 \\

\midrule
\multicolumn{13}{c}{\textbf{Qwen3.6-Plus-2026-04-02}} \\
4 & 0.697 & 0.683 & 0.690 & 0.942 & 0.947 & 0.944 & 0.783 & 0.786 & 0.784 & 0.367 & 0.317 & 0.342 \\
5 & 0.670 & 0.657 & 0.663 & 0.924 & 0.933 & 0.929 & 0.749 & 0.747 & 0.748 & 0.338 & 0.292 & 0.315 \\
6 & 0.657 & 0.645 & 0.651 & 0.909 & 0.919 & 0.914 & 0.733 & 0.733 & 0.733 & 0.330 & 0.281 & 0.305 \\
7 & 0.646 & 0.636 & 0.641 & 0.903 & 0.916 & 0.909 & 0.718 & 0.720 & 0.719 & 0.317 & 0.273 & 0.295 \\
8 & 0.640 & 0.633 & 0.637 & 0.896 & 0.914 & 0.905 & 0.715 & 0.715 & 0.715 & 0.309 & 0.269 & 0.289 \\
9 & 0.638 & 0.631 & 0.635 & 0.894 & 0.914 & 0.904 & 0.714 & 0.711 & 0.712 & 0.307 & 0.268 & 0.287 \\
10 & 0.636 & 0.630 & 0.633 & 0.889 & 0.912 & 0.901 & 0.712 & 0.708 & 0.710 & 0.306 & 0.270 & 0.288 \\

\midrule
\multicolumn{13}{c}{\textbf{Qwen-VL-Max}} \\
4 & 0.707 & 0.699 & 0.703 & 0.967 & 0.972 & 0.970 & 0.831 & 0.819 & 0.825 & 0.325 & 0.306 & 0.316 \\
5 & 0.697 & 0.691 & 0.694 & 0.963 & 0.969 & 0.966 & 0.812 & 0.808 & 0.810 & 0.317 & 0.296 & 0.306 \\
6 & 0.694 & 0.690 & 0.692 & 0.958 & 0.967 & 0.962 & 0.809 & 0.806 & 0.808 & 0.316 & 0.295 & 0.305 \\
7 & 0.692 & 0.686 & 0.689 & 0.958 & 0.965 & 0.962 & 0.803 & 0.799 & 0.801 & 0.315 & 0.293 & 0.304 \\
8 & 0.690 & 0.685 & 0.688 & 0.957 & 0.964 & 0.960 & 0.803 & 0.798 & 0.800 & 0.312 & 0.293 & 0.302 \\
9 & 0.690 & 0.685 & 0.688 & 0.956 & 0.964 & 0.960 & 0.800 & 0.796 & 0.798 & 0.312 & 0.296 & 0.304 \\
10 & 0.689 & 0.687 & 0.688 & 0.956 & 0.963 & 0.960 & 0.800 & 0.794 & 0.797 & 0.313 & 0.302 & 0.307 \\

\bottomrule
\caption{Full cumulative turn-indexed CorrTP@K results for all evaluated models on false-premise turns. O, Id, Attr, and Loc denote Overall, Identity, Attribute, and Location, respectively. GPT denotes the GPT-5.4 detector, Gemini denotes the Gemini-3.1-Pro-Preview detector, and Avg denotes the arithmetic mean of the two detector labels. We report only turns $K=4$ to $K=10$, because these are the false-premise turns in FPCO-Dialog.}
\label{tab:full_corrtp_numbers}
\end{longtable}

\endgroup
\begingroup
\small
\setlength{\tabcolsep}{2.2pt}
\renewcommand{\arraystretch}{1.04}

\begin{longtable}{@{}lrrrrrrrrrrrr@{}}
\toprule
K & O-GPT & O-Gem & O-Avg & Id-GPT & Id-Gem & Id-Avg & Attr-GPT & Attr-Gem & Attr-Avg & Loc-GPT & Loc-Gem & Loc-Avg \\
\midrule
\endfirsthead

\toprule
K & O-GPT & O-Gem & O-Avg & Id-GPT & Id-Gem & Id-Avg & Attr-GPT & Attr-Gem & Attr-Avg & Loc-GPT & Loc-Gem & Loc-Avg \\
\midrule
\endhead

\multicolumn{13}{c}{\textbf{Gemini 2.5 Flash}} \\
4 & 0.650 & 0.641 & 0.645 & 0.928 & 0.944 & 0.936 & 0.756 & 0.753 & 0.754 & 0.267 & 0.225 & 0.246 \\
5 & 0.672 & 0.649 & 0.661 & 0.964 & 0.967 & 0.966 & 0.781 & 0.769 & 0.775 & 0.272 & 0.211 & 0.241 \\
6 & 0.659 & 0.648 & 0.653 & 0.964 & 0.967 & 0.966 & 0.772 & 0.769 & 0.770 & 0.242 & 0.208 & 0.225 \\
7 & 0.651 & 0.638 & 0.645 & 0.956 & 0.956 & 0.956 & 0.767 & 0.761 & 0.764 & 0.231 & 0.197 & 0.214 \\
8 & 0.668 & 0.652 & 0.660 & 0.972 & 0.964 & 0.968 & 0.783 & 0.775 & 0.779 & 0.247 & 0.217 & 0.232 \\
9 & 0.662 & 0.648 & 0.655 & 0.958 & 0.956 & 0.957 & 0.781 & 0.772 & 0.776 & 0.247 & 0.217 & 0.232 \\
10 & 0.665 & 0.653 & 0.659 & 0.961 & 0.964 & 0.962 & 0.786 & 0.775 & 0.780 & 0.247 & 0.219 & 0.233 \\

\midrule
\multicolumn{13}{c}{\textbf{Gemini 2.5 Pro}} \\
4 & 0.704 & 0.688 & 0.696 & 0.947 & 0.947 & 0.947 & 0.811 & 0.800 & 0.806 & 0.353 & 0.317 & 0.335 \\
5 & 0.657 & 0.651 & 0.654 & 0.931 & 0.936 & 0.933 & 0.742 & 0.733 & 0.738 & 0.300 & 0.283 & 0.291 \\
6 & 0.656 & 0.646 & 0.651 & 0.931 & 0.931 & 0.931 & 0.744 & 0.744 & 0.744 & 0.294 & 0.264 & 0.279 \\
7 & 0.630 & 0.623 & 0.627 & 0.917 & 0.925 & 0.921 & 0.706 & 0.703 & 0.704 & 0.267 & 0.242 & 0.255 \\
8 & 0.638 & 0.640 & 0.639 & 0.917 & 0.922 & 0.919 & 0.717 & 0.728 & 0.722 & 0.281 & 0.269 & 0.275 \\
9 & 0.644 & 0.636 & 0.640 & 0.919 & 0.922 & 0.921 & 0.725 & 0.731 & 0.728 & 0.286 & 0.256 & 0.271 \\
10 & 0.640 & 0.638 & 0.639 & 0.917 & 0.922 & 0.919 & 0.725 & 0.714 & 0.720 & 0.278 & 0.278 & 0.278 \\

\midrule
\multicolumn{13}{c}{\textbf{Gemini 3.1 Pro Preview}} \\
4 & 0.606 & 0.601 & 0.603 & 0.808 & 0.822 & 0.815 & 0.647 & 0.658 & 0.653 & 0.361 & 0.322 & 0.342 \\
5 & 0.590 & 0.582 & 0.586 & 0.806 & 0.811 & 0.808 & 0.639 & 0.647 & 0.643 & 0.325 & 0.289 & 0.307 \\
6 & 0.592 & 0.586 & 0.589 & 0.792 & 0.800 & 0.796 & 0.647 & 0.650 & 0.649 & 0.336 & 0.308 & 0.322 \\
7 & 0.578 & 0.574 & 0.576 & 0.797 & 0.800 & 0.798 & 0.631 & 0.639 & 0.635 & 0.306 & 0.283 & 0.294 \\
8 & 0.588 & 0.582 & 0.585 & 0.797 & 0.803 & 0.800 & 0.647 & 0.650 & 0.649 & 0.319 & 0.294 & 0.306 \\
9 & 0.591 & 0.584 & 0.587 & 0.800 & 0.811 & 0.806 & 0.647 & 0.647 & 0.647 & 0.325 & 0.294 & 0.309 \\
10 & 0.591 & 0.588 & 0.589 & 0.808 & 0.811 & 0.810 & 0.642 & 0.647 & 0.645 & 0.322 & 0.306 & 0.314 \\

\midrule
\multicolumn{13}{c}{\textbf{GPT-4.1}} \\
4 & 0.619 & 0.611 & 0.615 & 0.917 & 0.917 & 0.917 & 0.689 & 0.678 & 0.683 & 0.253 & 0.239 & 0.246 \\
5 & 0.601 & 0.587 & 0.594 & 0.911 & 0.908 & 0.909 & 0.678 & 0.669 & 0.673 & 0.214 & 0.183 & 0.199 \\
6 & 0.592 & 0.585 & 0.589 & 0.903 & 0.908 & 0.905 & 0.661 & 0.661 & 0.661 & 0.211 & 0.186 & 0.199 \\
7 & 0.583 & 0.579 & 0.581 & 0.897 & 0.903 & 0.900 & 0.653 & 0.642 & 0.647 & 0.200 & 0.192 & 0.196 \\
8 & 0.581 & 0.569 & 0.575 & 0.892 & 0.900 & 0.896 & 0.650 & 0.636 & 0.643 & 0.200 & 0.172 & 0.186 \\
9 & 0.582 & 0.573 & 0.577 & 0.897 & 0.900 & 0.899 & 0.647 & 0.636 & 0.641 & 0.203 & 0.183 & 0.193 \\
10 & 0.587 & 0.573 & 0.580 & 0.889 & 0.903 & 0.896 & 0.658 & 0.633 & 0.645 & 0.214 & 0.183 & 0.199 \\

\midrule
\multicolumn{13}{c}{\textbf{GPT-4o}} \\
4 & 0.523 & 0.490 & 0.506 & 0.878 & 0.872 & 0.875 & 0.489 & 0.467 & 0.478 & 0.203 & 0.131 & 0.167 \\
5 & 0.500 & 0.461 & 0.481 & 0.850 & 0.853 & 0.851 & 0.467 & 0.456 & 0.462 & 0.183 & 0.075 & 0.129 \\
6 & 0.501 & 0.461 & 0.481 & 0.853 & 0.850 & 0.851 & 0.467 & 0.456 & 0.462 & 0.183 & 0.078 & 0.131 \\
7 & 0.495 & 0.456 & 0.476 & 0.850 & 0.850 & 0.850 & 0.469 & 0.453 & 0.461 & 0.167 & 0.064 & 0.116 \\
8 & 0.499 & 0.457 & 0.478 & 0.853 & 0.850 & 0.851 & 0.467 & 0.453 & 0.460 & 0.178 & 0.069 & 0.123 \\
9 & 0.499 & 0.456 & 0.478 & 0.858 & 0.856 & 0.857 & 0.467 & 0.450 & 0.459 & 0.172 & 0.061 & 0.116 \\
10 & 0.494 & 0.456 & 0.475 & 0.842 & 0.844 & 0.843 & 0.472 & 0.461 & 0.467 & 0.169 & 0.061 & 0.115 \\

\midrule
\multicolumn{13}{c}{\textbf{GPT-5}} \\
4 & 0.072 & 0.075 & 0.073 & 0.125 & 0.133 & 0.129 & 0.069 & 0.083 & 0.076 & 0.022 & 0.008 & 0.015 \\
5 & 0.069 & 0.077 & 0.073 & 0.117 & 0.128 & 0.122 & 0.083 & 0.097 & 0.090 & 0.008 & 0.006 & 0.007 \\
6 & 0.075 & 0.077 & 0.076 & 0.136 & 0.133 & 0.135 & 0.075 & 0.089 & 0.082 & 0.014 & 0.008 & 0.011 \\
7 & 0.081 & 0.086 & 0.083 & 0.144 & 0.150 & 0.147 & 0.089 & 0.103 & 0.096 & 0.008 & 0.006 & 0.007 \\
8 & 0.080 & 0.081 & 0.081 & 0.139 & 0.142 & 0.141 & 0.089 & 0.100 & 0.095 & 0.011 & 0.003 & 0.007 \\
9 & 0.081 & 0.085 & 0.083 & 0.144 & 0.147 & 0.145 & 0.092 & 0.108 & 0.100 & 0.006 & 0.000 & 0.003 \\
10 & 0.081 & 0.084 & 0.083 & 0.150 & 0.153 & 0.151 & 0.083 & 0.097 & 0.090 & 0.008 & 0.003 & 0.005 \\

\midrule
\multicolumn{13}{c}{\textbf{GPT-5.4}} \\
4 & 0.376 & 0.360 & 0.368 & 0.575 & 0.542 & 0.558 & 0.433 & 0.425 & 0.429 & 0.119 & 0.114 & 0.116 \\
5 & 0.326 & 0.318 & 0.322 & 0.492 & 0.475 & 0.483 & 0.389 & 0.383 & 0.386 & 0.097 & 0.094 & 0.096 \\
6 & 0.295 & 0.281 & 0.288 & 0.406 & 0.392 & 0.399 & 0.369 & 0.364 & 0.366 & 0.111 & 0.086 & 0.099 \\
7 & 0.280 & 0.269 & 0.275 & 0.375 & 0.361 & 0.368 & 0.367 & 0.364 & 0.365 & 0.097 & 0.083 & 0.090 \\
8 & 0.277 & 0.267 & 0.272 & 0.383 & 0.369 & 0.376 & 0.364 & 0.361 & 0.362 & 0.083 & 0.069 & 0.076 \\
9 & 0.269 & 0.263 & 0.266 & 0.358 & 0.358 & 0.358 & 0.367 & 0.356 & 0.361 & 0.083 & 0.075 & 0.079 \\
10 & 0.271 & 0.263 & 0.267 & 0.364 & 0.364 & 0.364 & 0.353 & 0.353 & 0.353 & 0.097 & 0.072 & 0.084 \\

\midrule
\multicolumn{13}{c}{\textbf{InternVL2.5-2B}} \\
4 & 0.139 & 0.131 & 0.135 & 0.178 & 0.175 & 0.176 & 0.094 & 0.092 & 0.093 & 0.144 & 0.128 & 0.136 \\
5 & 0.058 & 0.054 & 0.056 & 0.097 & 0.086 & 0.091 & 0.022 & 0.025 & 0.024 & 0.056 & 0.050 & 0.053 \\
6 & 0.040 & 0.037 & 0.038 & 0.069 & 0.067 & 0.068 & 0.019 & 0.022 & 0.020 & 0.031 & 0.022 & 0.026 \\
7 & 0.039 & 0.032 & 0.036 & 0.053 & 0.050 & 0.052 & 0.042 & 0.033 & 0.038 & 0.022 & 0.014 & 0.018 \\
8 & 0.036 & 0.031 & 0.034 & 0.058 & 0.053 & 0.056 & 0.019 & 0.019 & 0.019 & 0.031 & 0.019 & 0.025 \\
9 & 0.034 & 0.029 & 0.032 & 0.053 & 0.050 & 0.052 & 0.031 & 0.025 & 0.028 & 0.019 & 0.011 & 0.015 \\
10 & 0.031 & 0.029 & 0.030 & 0.044 & 0.042 & 0.043 & 0.017 & 0.022 & 0.019 & 0.031 & 0.022 & 0.026 \\

\midrule
\multicolumn{13}{c}{\textbf{InternVL2.5-4B}} \\
4 & 0.145 & 0.134 & 0.140 & 0.217 & 0.211 & 0.214 & 0.106 & 0.103 & 0.104 & 0.114 & 0.089 & 0.102 \\
5 & 0.075 & 0.070 & 0.073 & 0.153 & 0.147 & 0.150 & 0.042 & 0.036 & 0.039 & 0.031 & 0.028 & 0.029 \\
6 & 0.067 & 0.065 & 0.066 & 0.139 & 0.142 & 0.141 & 0.036 & 0.036 & 0.036 & 0.025 & 0.017 & 0.021 \\
7 & 0.059 & 0.059 & 0.059 & 0.131 & 0.133 & 0.132 & 0.033 & 0.033 & 0.033 & 0.014 & 0.011 & 0.013 \\
8 & 0.057 & 0.056 & 0.057 & 0.125 & 0.128 & 0.127 & 0.031 & 0.025 & 0.028 & 0.017 & 0.014 & 0.015 \\
9 & 0.066 & 0.064 & 0.065 & 0.136 & 0.136 & 0.136 & 0.036 & 0.036 & 0.036 & 0.025 & 0.019 & 0.022 \\
10 & 0.064 & 0.061 & 0.062 & 0.139 & 0.136 & 0.138 & 0.033 & 0.031 & 0.032 & 0.019 & 0.017 & 0.018 \\

\midrule
\multicolumn{13}{c}{\textbf{InternVL2.5-8B}} \\
4 & 0.152 & 0.147 & 0.149 & 0.283 & 0.281 & 0.282 & 0.103 & 0.100 & 0.102 & 0.069 & 0.061 & 0.065 \\
5 & 0.111 & 0.106 & 0.108 & 0.217 & 0.219 & 0.218 & 0.075 & 0.072 & 0.073 & 0.042 & 0.025 & 0.034 \\
6 & 0.105 & 0.101 & 0.103 & 0.222 & 0.219 & 0.221 & 0.069 & 0.067 & 0.068 & 0.022 & 0.017 & 0.019 \\
7 & 0.101 & 0.098 & 0.100 & 0.214 & 0.214 & 0.214 & 0.072 & 0.069 & 0.071 & 0.017 & 0.011 & 0.014 \\
8 & 0.103 & 0.102 & 0.102 & 0.219 & 0.219 & 0.219 & 0.069 & 0.069 & 0.069 & 0.019 & 0.017 & 0.018 \\
9 & 0.104 & 0.100 & 0.102 & 0.217 & 0.217 & 0.217 & 0.075 & 0.072 & 0.073 & 0.019 & 0.011 & 0.015 \\
10 & 0.108 & 0.105 & 0.106 & 0.217 & 0.217 & 0.217 & 0.089 & 0.086 & 0.087 & 0.019 & 0.011 & 0.015 \\

\midrule
\multicolumn{13}{c}{\textbf{LLaVA-NeXT-13B}} \\
4 & 0.165 & 0.148 & 0.157 & 0.292 & 0.289 & 0.290 & 0.094 & 0.083 & 0.088 & 0.108 & 0.072 & 0.090 \\
5 & 0.115 & 0.092 & 0.104 & 0.203 & 0.194 & 0.199 & 0.053 & 0.047 & 0.050 & 0.089 & 0.033 & 0.061 \\
6 & 0.093 & 0.078 & 0.085 & 0.156 & 0.158 & 0.157 & 0.047 & 0.036 & 0.041 & 0.075 & 0.039 & 0.057 \\
7 & 0.079 & 0.074 & 0.076 & 0.150 & 0.153 & 0.151 & 0.050 & 0.044 & 0.047 & 0.036 & 0.025 & 0.030 \\
8 & 0.083 & 0.071 & 0.077 & 0.150 & 0.147 & 0.148 & 0.050 & 0.039 & 0.044 & 0.050 & 0.028 & 0.039 \\
9 & 0.081 & 0.069 & 0.075 & 0.142 & 0.136 & 0.139 & 0.042 & 0.042 & 0.042 & 0.058 & 0.028 & 0.043 \\
10 & 0.081 & 0.067 & 0.074 & 0.144 & 0.142 & 0.143 & 0.039 & 0.031 & 0.035 & 0.058 & 0.028 & 0.043 \\

\midrule
\multicolumn{13}{c}{\textbf{LLaVA-OneVision-0.5B}} \\
4 & 0.060 & 0.051 & 0.055 & 0.047 & 0.053 & 0.050 & 0.025 & 0.039 & 0.032 & 0.108 & 0.061 & 0.084 \\
5 & 0.029 & 0.014 & 0.022 & 0.011 & 0.003 & 0.007 & 0.014 & 0.014 & 0.014 & 0.061 & 0.025 & 0.043 \\
6 & 0.027 & 0.016 & 0.021 & 0.008 & 0.006 & 0.007 & 0.008 & 0.011 & 0.009 & 0.064 & 0.031 & 0.048 \\
7 & 0.024 & 0.008 & 0.016 & 0.008 & 0.006 & 0.007 & 0.014 & 0.011 & 0.013 & 0.050 & 0.008 & 0.029 \\
8 & 0.020 & 0.008 & 0.014 & 0.003 & 0.003 & 0.003 & 0.014 & 0.014 & 0.014 & 0.044 & 0.008 & 0.026 \\
9 & 0.020 & 0.010 & 0.015 & 0.008 & 0.008 & 0.008 & 0.014 & 0.014 & 0.014 & 0.039 & 0.008 & 0.024 \\
10 & 0.015 & 0.006 & 0.010 & 0.008 & 0.006 & 0.007 & 0.003 & 0.003 & 0.003 & 0.033 & 0.011 & 0.022 \\

\midrule
\multicolumn{13}{c}{\textbf{LLaVA-OneVision-7B}} \\
4 & 0.021 & 0.031 & 0.026 & 0.019 & 0.042 & 0.030 & 0.025 & 0.039 & 0.032 & 0.019 & 0.014 & 0.017 \\
5 & 0.011 & 0.010 & 0.010 & 0.014 & 0.008 & 0.011 & 0.011 & 0.014 & 0.013 & 0.008 & 0.008 & 0.008 \\
6 & 0.008 & 0.008 & 0.008 & 0.003 & 0.006 & 0.005 & 0.014 & 0.017 & 0.015 & 0.008 & 0.003 & 0.005 \\
7 & 0.006 & 0.005 & 0.005 & 0.003 & 0.003 & 0.003 & 0.006 & 0.008 & 0.007 & 0.008 & 0.003 & 0.005 \\
8 & 0.011 & 0.005 & 0.008 & 0.003 & 0.003 & 0.003 & 0.008 & 0.008 & 0.008 & 0.022 & 0.003 & 0.012 \\
9 & 0.006 & 0.006 & 0.006 & 0.003 & 0.006 & 0.005 & 0.008 & 0.011 & 0.009 & 0.008 & 0.003 & 0.005 \\
10 & 0.005 & 0.004 & 0.005 & 0.006 & 0.003 & 0.005 & 0.003 & 0.008 & 0.005 & 0.006 & 0.000 & 0.003 \\

\midrule
\multicolumn{13}{c}{\textbf{Qwen2.5-VL-32B-Instruct}} \\
4 & 0.629 & 0.620 & 0.625 & 0.894 & 0.900 & 0.897 & 0.717 & 0.700 & 0.708 & 0.275 & 0.261 & 0.268 \\
5 & 0.573 & 0.574 & 0.573 & 0.847 & 0.853 & 0.850 & 0.661 & 0.667 & 0.664 & 0.211 & 0.203 & 0.207 \\
6 & 0.569 & 0.576 & 0.573 & 0.864 & 0.869 & 0.867 & 0.653 & 0.667 & 0.660 & 0.192 & 0.192 & 0.192 \\
7 & 0.556 & 0.568 & 0.562 & 0.847 & 0.858 & 0.853 & 0.642 & 0.653 & 0.647 & 0.178 & 0.192 & 0.185 \\
8 & 0.553 & 0.559 & 0.556 & 0.858 & 0.869 & 0.863 & 0.647 & 0.642 & 0.645 & 0.153 & 0.167 & 0.160 \\
9 & 0.554 & 0.550 & 0.552 & 0.850 & 0.844 & 0.847 & 0.636 & 0.642 & 0.639 & 0.175 & 0.164 & 0.169 \\
10 & 0.554 & 0.554 & 0.554 & 0.856 & 0.872 & 0.864 & 0.650 & 0.631 & 0.641 & 0.156 & 0.158 & 0.157 \\

\midrule
\multicolumn{13}{c}{\textbf{Qwen2.5-VL-3B-Instruct}} \\
4 & 0.316 & 0.283 & 0.299 & 0.425 & 0.417 & 0.421 & 0.328 & 0.303 & 0.316 & 0.194 & 0.131 & 0.163 \\
5 & 0.184 & 0.174 & 0.179 & 0.300 & 0.308 & 0.304 & 0.161 & 0.153 & 0.157 & 0.092 & 0.061 & 0.076 \\
6 & 0.153 & 0.146 & 0.149 & 0.269 & 0.272 & 0.271 & 0.136 & 0.131 & 0.134 & 0.053 & 0.036 & 0.044 \\
7 & 0.144 & 0.134 & 0.139 & 0.258 & 0.253 & 0.256 & 0.136 & 0.128 & 0.132 & 0.036 & 0.022 & 0.029 \\
8 & 0.147 & 0.138 & 0.143 & 0.256 & 0.256 & 0.256 & 0.147 & 0.139 & 0.143 & 0.039 & 0.019 & 0.029 \\
9 & 0.138 & 0.130 & 0.134 & 0.247 & 0.250 & 0.248 & 0.122 & 0.119 & 0.120 & 0.044 & 0.019 & 0.032 \\
10 & 0.144 & 0.136 & 0.140 & 0.250 & 0.244 & 0.247 & 0.139 & 0.142 & 0.141 & 0.044 & 0.022 & 0.033 \\

\midrule
\multicolumn{13}{c}{\textbf{Qwen2.5-VL-72B-Instruct}} \\
4 & 0.647 & 0.622 & 0.635 & 0.900 & 0.900 & 0.900 & 0.747 & 0.733 & 0.740 & 0.294 & 0.233 & 0.264 \\
5 & 0.622 & 0.606 & 0.614 & 0.900 & 0.900 & 0.900 & 0.736 & 0.728 & 0.732 & 0.231 & 0.192 & 0.212 \\
6 & 0.641 & 0.618 & 0.629 & 0.900 & 0.900 & 0.900 & 0.778 & 0.764 & 0.771 & 0.244 & 0.189 & 0.216 \\
7 & 0.628 & 0.611 & 0.619 & 0.903 & 0.900 & 0.901 & 0.756 & 0.744 & 0.750 & 0.225 & 0.189 & 0.207 \\
8 & 0.639 & 0.619 & 0.629 & 0.914 & 0.914 & 0.914 & 0.761 & 0.750 & 0.756 & 0.242 & 0.192 & 0.217 \\
9 & 0.641 & 0.619 & 0.630 & 0.908 & 0.911 & 0.909 & 0.772 & 0.753 & 0.762 & 0.242 & 0.192 & 0.217 \\
10 & 0.656 & 0.634 & 0.645 & 0.919 & 0.917 & 0.918 & 0.781 & 0.769 & 0.775 & 0.269 & 0.217 & 0.243 \\

\midrule
\multicolumn{13}{c}{\textbf{Qwen2.5-VL-7B-Instruct}} \\
4 & 0.610 & 0.563 & 0.587 & 0.919 & 0.906 & 0.913 & 0.625 & 0.594 & 0.609 & 0.286 & 0.189 & 0.237 \\
5 & 0.510 & 0.492 & 0.501 & 0.836 & 0.833 & 0.835 & 0.564 & 0.550 & 0.557 & 0.131 & 0.092 & 0.112 \\
6 & 0.480 & 0.465 & 0.473 & 0.800 & 0.803 & 0.802 & 0.522 & 0.506 & 0.514 & 0.117 & 0.086 & 0.102 \\
7 & 0.465 & 0.459 & 0.462 & 0.794 & 0.806 & 0.800 & 0.517 & 0.508 & 0.512 & 0.083 & 0.064 & 0.074 \\
8 & 0.467 & 0.459 & 0.463 & 0.811 & 0.814 & 0.812 & 0.511 & 0.500 & 0.506 & 0.078 & 0.064 & 0.071 \\
9 & 0.470 & 0.461 & 0.466 & 0.786 & 0.811 & 0.798 & 0.522 & 0.508 & 0.515 & 0.103 & 0.064 & 0.083 \\
10 & 0.477 & 0.461 & 0.469 & 0.811 & 0.811 & 0.811 & 0.533 & 0.514 & 0.524 & 0.086 & 0.058 & 0.072 \\

\midrule
\multicolumn{13}{c}{\textbf{Qwen3.5-Plus-2026-04-20}} \\
4 & 0.680 & 0.661 & 0.671 & 0.933 & 0.939 & 0.936 & 0.758 & 0.744 & 0.751 & 0.347 & 0.300 & 0.324 \\
5 & 0.623 & 0.603 & 0.613 & 0.903 & 0.911 & 0.907 & 0.669 & 0.650 & 0.659 & 0.297 & 0.247 & 0.272 \\
6 & 0.614 & 0.601 & 0.607 & 0.869 & 0.892 & 0.881 & 0.661 & 0.653 & 0.657 & 0.311 & 0.258 & 0.284 \\
7 & 0.606 & 0.590 & 0.598 & 0.881 & 0.883 & 0.882 & 0.658 & 0.644 & 0.651 & 0.281 & 0.242 & 0.262 \\
8 & 0.615 & 0.601 & 0.608 & 0.881 & 0.903 & 0.892 & 0.686 & 0.667 & 0.677 & 0.278 & 0.233 & 0.256 \\
9 & 0.613 & 0.604 & 0.609 & 0.872 & 0.892 & 0.882 & 0.669 & 0.656 & 0.663 & 0.297 & 0.264 & 0.280 \\
10 & 0.616 & 0.616 & 0.616 & 0.881 & 0.914 & 0.897 & 0.675 & 0.656 & 0.665 & 0.292 & 0.278 & 0.285 \\

\midrule
\multicolumn{13}{c}{\textbf{Qwen3.6-Plus-2026-04-02}} \\
4 & 0.697 & 0.683 & 0.690 & 0.942 & 0.947 & 0.944 & 0.783 & 0.786 & 0.784 & 0.367 & 0.317 & 0.342 \\
5 & 0.643 & 0.631 & 0.637 & 0.906 & 0.919 & 0.913 & 0.714 & 0.708 & 0.711 & 0.308 & 0.267 & 0.287 \\
6 & 0.632 & 0.619 & 0.625 & 0.881 & 0.892 & 0.887 & 0.703 & 0.706 & 0.704 & 0.314 & 0.261 & 0.287 \\
7 & 0.611 & 0.611 & 0.611 & 0.883 & 0.906 & 0.895 & 0.672 & 0.681 & 0.677 & 0.278 & 0.247 & 0.263 \\
8 & 0.618 & 0.619 & 0.619 & 0.869 & 0.908 & 0.889 & 0.703 & 0.694 & 0.698 & 0.281 & 0.256 & 0.269 \\
9 & 0.630 & 0.621 & 0.625 & 0.883 & 0.914 & 0.899 & 0.708 & 0.689 & 0.698 & 0.297 & 0.261 & 0.279 \\
10 & 0.619 & 0.624 & 0.621 & 0.858 & 0.900 & 0.879 & 0.697 & 0.689 & 0.693 & 0.300 & 0.283 & 0.291 \\

\midrule
\multicolumn{13}{c}{\textbf{Qwen-VL-Max}} \\
4 & 0.707 & 0.699 & 0.703 & 0.967 & 0.972 & 0.970 & 0.831 & 0.819 & 0.825 & 0.325 & 0.306 & 0.316 \\
5 & 0.687 & 0.683 & 0.685 & 0.958 & 0.967 & 0.962 & 0.794 & 0.797 & 0.796 & 0.308 & 0.286 & 0.297 \\
6 & 0.689 & 0.686 & 0.688 & 0.950 & 0.961 & 0.956 & 0.803 & 0.803 & 0.803 & 0.314 & 0.294 & 0.304 \\
7 & 0.683 & 0.674 & 0.679 & 0.956 & 0.961 & 0.958 & 0.783 & 0.775 & 0.779 & 0.311 & 0.286 & 0.298 \\
8 & 0.685 & 0.683 & 0.684 & 0.953 & 0.961 & 0.957 & 0.803 & 0.797 & 0.800 & 0.300 & 0.292 & 0.296 \\
9 & 0.685 & 0.686 & 0.685 & 0.953 & 0.961 & 0.957 & 0.789 & 0.783 & 0.786 & 0.314 & 0.314 & 0.314 \\
10 & 0.689 & 0.694 & 0.692 & 0.956 & 0.961 & 0.958 & 0.794 & 0.786 & 0.790 & 0.317 & 0.333 & 0.325 \\

\bottomrule
\caption{Full turn-wise TurnCorr@K results for all evaluated models on false-premise turns. O, Id, Attr, and Loc denote Overall, Identity, Attribute, and Location, respectively. GPT denotes the GPT-5.4 detector, Gemini denotes the Gemini-3.1-Pro-Preview detector, and Avg denotes the arithmetic mean of the two detector labels. We report only turns $K=4$ to $K=10$, because these are the false-premise turns in FPCO-Dialog.}
\label{tab:full_turncorr_numbers}
\end{longtable}

\endgroup
\begingroup
\small
\setlength{\tabcolsep}{2.2pt}
\renewcommand{\arraystretch}{1.04}

\begin{longtable}{@{}lrrrrrrrrrrrr@{}}
\toprule
K & O-GPT & O-Gem & O-Avg & Id-GPT & Id-Gem & Id-Avg & Attr-GPT & Attr-Gem & Attr-Avg & Loc-GPT & Loc-Gem & Loc-Avg \\
\midrule
\endfirsthead

\toprule
K & O-GPT & O-Gem & O-Avg & Id-GPT & Id-Gem & Id-Avg & Attr-GPT & Attr-Gem & Attr-Avg & Loc-GPT & Loc-Gem & Loc-Avg \\
\midrule
\endhead

\multicolumn{13}{c}{\textbf{Gemini 2.5 Flash}} \\
1 & 0.020 & 0.018 & 0.019 & 0.017 & 0.011 & 0.014 & 0.036 & 0.033 & 0.035 & 0.008 & 0.008 & 0.008 \\
2 & 0.020 & 0.017 & 0.019 & 0.015 & 0.011 & 0.013 & 0.033 & 0.032 & 0.033 & 0.011 & 0.008 & 0.009 \\
3 & 0.019 & 0.017 & 0.018 & 0.014 & 0.012 & 0.013 & 0.033 & 0.031 & 0.032 & 0.009 & 0.007 & 0.008 \\

\midrule
\multicolumn{13}{c}{\textbf{Gemini 2.5 Pro}} \\
1 & 0.026 & 0.019 & 0.022 & 0.017 & 0.011 & 0.014 & 0.039 & 0.031 & 0.035 & 0.022 & 0.014 & 0.018 \\
2 & 0.023 & 0.015 & 0.019 & 0.017 & 0.014 & 0.015 & 0.032 & 0.024 & 0.028 & 0.019 & 0.008 & 0.013 \\
3 & 0.020 & 0.013 & 0.017 & 0.016 & 0.011 & 0.013 & 0.028 & 0.020 & 0.024 & 0.017 & 0.006 & 0.011 \\

\midrule
\multicolumn{13}{c}{\textbf{Gemini 3.1 Pro Preview}} \\
1 & 0.031 & 0.026 & 0.028 & 0.025 & 0.022 & 0.024 & 0.042 & 0.033 & 0.038 & 0.025 & 0.022 & 0.024 \\
2 & 0.025 & 0.021 & 0.023 & 0.018 & 0.017 & 0.018 & 0.036 & 0.029 & 0.033 & 0.021 & 0.017 & 0.019 \\
3 & 0.022 & 0.019 & 0.020 & 0.017 & 0.015 & 0.016 & 0.032 & 0.028 & 0.030 & 0.018 & 0.014 & 0.016 \\

\midrule
\multicolumn{13}{c}{\textbf{GPT-4.1}} \\
1 & 0.012 & 0.009 & 0.010 & 0.003 & 0.003 & 0.003 & 0.019 & 0.011 & 0.015 & 0.014 & 0.014 & 0.014 \\
2 & 0.011 & 0.008 & 0.009 & 0.003 & 0.003 & 0.003 & 0.015 & 0.008 & 0.011 & 0.015 & 0.014 & 0.014 \\
3 & 0.010 & 0.008 & 0.009 & 0.003 & 0.003 & 0.003 & 0.013 & 0.007 & 0.010 & 0.014 & 0.013 & 0.013 \\

\midrule
\multicolumn{13}{c}{\textbf{GPT-4o}} \\
1 & 0.010 & 0.008 & 0.009 & 0.008 & 0.006 & 0.007 & 0.017 & 0.014 & 0.015 & 0.006 & 0.006 & 0.006 \\
2 & 0.009 & 0.008 & 0.009 & 0.007 & 0.006 & 0.007 & 0.015 & 0.015 & 0.015 & 0.006 & 0.003 & 0.005 \\
3 & 0.008 & 0.007 & 0.007 & 0.006 & 0.005 & 0.005 & 0.015 & 0.014 & 0.014 & 0.005 & 0.002 & 0.004 \\

\midrule
\multicolumn{13}{c}{\textbf{GPT-5}} \\
1 & 0.000 & 0.000 & 0.000 & 0.000 & 0.000 & 0.000 & 0.000 & 0.000 & 0.000 & 0.000 & 0.000 & 0.000 \\
2 & 0.000 & 0.000 & 0.000 & 0.000 & 0.000 & 0.000 & 0.000 & 0.000 & 0.000 & 0.000 & 0.000 & 0.000 \\
3 & 0.000 & 0.000 & 0.000 & 0.000 & 0.000 & 0.000 & 0.000 & 0.000 & 0.000 & 0.000 & 0.000 & 0.000 \\

\midrule
\multicolumn{13}{c}{\textbf{GPT-5.4}} \\
1 & 0.001 & 0.000 & 0.001 & 0.000 & 0.000 & 0.000 & 0.003 & 0.000 & 0.002 & 0.000 & 0.000 & 0.000 \\
2 & 0.001 & 0.000 & 0.001 & 0.000 & 0.000 & 0.000 & 0.001 & 0.000 & 0.001 & 0.001 & 0.000 & 0.001 \\
3 & 0.001 & 0.000 & 0.001 & 0.000 & 0.000 & 0.000 & 0.001 & 0.000 & 0.001 & 0.001 & 0.000 & 0.001 \\

\midrule
\multicolumn{13}{c}{\textbf{InternVL2.5-2B}} \\
1 & 0.015 & 0.009 & 0.012 & 0.006 & 0.006 & 0.006 & 0.025 & 0.014 & 0.019 & 0.014 & 0.008 & 0.011 \\
2 & 0.010 & 0.007 & 0.009 & 0.007 & 0.006 & 0.007 & 0.015 & 0.011 & 0.013 & 0.008 & 0.004 & 0.006 \\
3 & 0.009 & 0.005 & 0.007 & 0.005 & 0.004 & 0.005 & 0.014 & 0.009 & 0.011 & 0.009 & 0.003 & 0.006 \\

\midrule
\multicolumn{13}{c}{\textbf{InternVL2.5-4B}} \\
1 & 0.005 & 0.004 & 0.005 & 0.006 & 0.003 & 0.005 & 0.006 & 0.006 & 0.006 & 0.003 & 0.003 & 0.003 \\
2 & 0.004 & 0.003 & 0.004 & 0.004 & 0.003 & 0.004 & 0.007 & 0.006 & 0.007 & 0.001 & 0.001 & 0.001 \\
3 & 0.003 & 0.003 & 0.003 & 0.004 & 0.003 & 0.004 & 0.006 & 0.005 & 0.005 & 0.001 & 0.001 & 0.001 \\

\midrule
\multicolumn{13}{c}{\textbf{InternVL2.5-8B}} \\
1 & 0.006 & 0.003 & 0.005 & 0.006 & 0.000 & 0.003 & 0.011 & 0.006 & 0.009 & 0.003 & 0.003 & 0.003 \\
2 & 0.005 & 0.002 & 0.004 & 0.003 & 0.000 & 0.002 & 0.008 & 0.006 & 0.007 & 0.003 & 0.001 & 0.002 \\
3 & 0.005 & 0.002 & 0.004 & 0.002 & 0.000 & 0.001 & 0.007 & 0.006 & 0.007 & 0.005 & 0.001 & 0.003 \\

\midrule
\multicolumn{13}{c}{\textbf{LLaVA-NeXT-13B}} \\
1 & 0.006 & 0.005 & 0.005 & 0.003 & 0.003 & 0.003 & 0.008 & 0.006 & 0.007 & 0.006 & 0.006 & 0.006 \\
2 & 0.003 & 0.003 & 0.003 & 0.001 & 0.001 & 0.001 & 0.006 & 0.004 & 0.005 & 0.003 & 0.003 & 0.003 \\
3 & 0.002 & 0.002 & 0.002 & 0.001 & 0.001 & 0.001 & 0.005 & 0.004 & 0.005 & 0.002 & 0.002 & 0.002 \\

\midrule
\multicolumn{13}{c}{\textbf{LLaVA-OneVision-0.5B}} \\
1 & 0.004 & 0.001 & 0.003 & 0.003 & 0.000 & 0.002 & 0.006 & 0.003 & 0.005 & 0.003 & 0.000 & 0.002 \\
2 & 0.003 & 0.000 & 0.002 & 0.003 & 0.000 & 0.002 & 0.004 & 0.001 & 0.003 & 0.001 & 0.000 & 0.001 \\
3 & 0.002 & 0.000 & 0.001 & 0.003 & 0.000 & 0.002 & 0.003 & 0.001 & 0.002 & 0.002 & 0.000 & 0.001 \\

\midrule
\multicolumn{13}{c}{\textbf{LLaVA-OneVision-7B}} \\
1 & 0.000 & 0.000 & 0.000 & 0.000 & 0.000 & 0.000 & 0.000 & 0.000 & 0.000 & 0.000 & 0.000 & 0.000 \\
2 & 0.000 & 0.000 & 0.000 & 0.000 & 0.000 & 0.000 & 0.000 & 0.000 & 0.000 & 0.000 & 0.000 & 0.000 \\
3 & 0.000 & 0.000 & 0.000 & 0.000 & 0.000 & 0.000 & 0.000 & 0.000 & 0.000 & 0.000 & 0.000 & 0.000 \\

\midrule
\multicolumn{13}{c}{\textbf{Qwen2.5-VL-32B-Instruct}} \\
1 & 0.033 & 0.028 & 0.030 & 0.031 & 0.028 & 0.029 & 0.031 & 0.028 & 0.029 & 0.039 & 0.028 & 0.034 \\
2 & 0.029 & 0.024 & 0.027 & 0.022 & 0.019 & 0.020 & 0.032 & 0.028 & 0.030 & 0.033 & 0.025 & 0.029 \\
3 & 0.029 & 0.024 & 0.027 & 0.021 & 0.021 & 0.021 & 0.031 & 0.029 & 0.030 & 0.033 & 0.023 & 0.028 \\

\midrule
\multicolumn{13}{c}{\textbf{Qwen2.5-VL-3B-Instruct}} \\
1 & 0.006 & 0.002 & 0.004 & 0.006 & 0.003 & 0.005 & 0.006 & 0.003 & 0.005 & 0.006 & 0.000 & 0.003 \\
2 & 0.003 & 0.001 & 0.002 & 0.003 & 0.001 & 0.002 & 0.003 & 0.001 & 0.002 & 0.003 & 0.000 & 0.002 \\
3 & 0.003 & 0.001 & 0.002 & 0.003 & 0.001 & 0.002 & 0.003 & 0.001 & 0.002 & 0.003 & 0.000 & 0.002 \\

\midrule
\multicolumn{13}{c}{\textbf{Qwen2.5-VL-72B-Instruct}} \\
1 & 0.028 & 0.024 & 0.026 & 0.025 & 0.022 & 0.024 & 0.036 & 0.031 & 0.034 & 0.022 & 0.019 & 0.020 \\
2 & 0.024 & 0.022 & 0.023 & 0.018 & 0.017 & 0.018 & 0.033 & 0.031 & 0.032 & 0.021 & 0.019 & 0.020 \\
3 & 0.023 & 0.023 & 0.023 & 0.018 & 0.017 & 0.018 & 0.032 & 0.031 & 0.032 & 0.020 & 0.019 & 0.019 \\

\midrule
\multicolumn{13}{c}{\textbf{Qwen2.5-VL-7B-Instruct}} \\
1 & 0.020 & 0.016 & 0.018 & 0.025 & 0.019 & 0.022 & 0.025 & 0.019 & 0.022 & 0.011 & 0.008 & 0.009 \\
2 & 0.018 & 0.015 & 0.017 & 0.018 & 0.014 & 0.016 & 0.025 & 0.021 & 0.023 & 0.011 & 0.010 & 0.010 \\
3 & 0.017 & 0.014 & 0.015 & 0.016 & 0.012 & 0.014 & 0.025 & 0.021 & 0.023 & 0.010 & 0.009 & 0.009 \\

\midrule
\multicolumn{13}{c}{\textbf{Qwen3.5-Plus-2026-04-20}} \\
1 & 0.024 & 0.020 & 0.022 & 0.014 & 0.014 & 0.014 & 0.036 & 0.025 & 0.030 & 0.022 & 0.022 & 0.022 \\
2 & 0.021 & 0.018 & 0.019 & 0.014 & 0.014 & 0.014 & 0.029 & 0.021 & 0.025 & 0.019 & 0.018 & 0.018 \\
3 & 0.016 & 0.014 & 0.015 & 0.010 & 0.010 & 0.010 & 0.023 & 0.018 & 0.020 & 0.016 & 0.014 & 0.015 \\

\midrule
\multicolumn{13}{c}{\textbf{Qwen3.6-Plus-2026-04-02}} \\
1 & 0.017 & 0.017 & 0.017 & 0.006 & 0.011 & 0.009 & 0.025 & 0.028 & 0.027 & 0.019 & 0.011 & 0.015 \\
2 & 0.016 & 0.016 & 0.016 & 0.010 & 0.011 & 0.010 & 0.025 & 0.028 & 0.027 & 0.014 & 0.008 & 0.011 \\
3 & 0.015 & 0.015 & 0.015 & 0.009 & 0.010 & 0.009 & 0.024 & 0.024 & 0.024 & 0.012 & 0.009 & 0.010 \\

\midrule
\multicolumn{13}{c}{\textbf{Qwen-VL-Max}} \\
1 & 0.017 & 0.016 & 0.017 & 0.014 & 0.014 & 0.014 & 0.019 & 0.019 & 0.019 & 0.017 & 0.014 & 0.015 \\
2 & 0.017 & 0.017 & 0.017 & 0.013 & 0.014 & 0.013 & 0.019 & 0.019 & 0.019 & 0.018 & 0.017 & 0.018 \\
3 & 0.017 & 0.016 & 0.017 & 0.013 & 0.014 & 0.013 & 0.019 & 0.019 & 0.019 & 0.019 & 0.016 & 0.018 \\

\bottomrule
\caption{Full cumulative CorrFP@K results for all evaluated models on premise-correct turns. O denotes Overall, while Id, Attr, and Loc denote dialogues assigned to the Identity, Attribute, and Location false-premise classes, respectively. GPT denotes the GPT-5.4 detector, Gemini denotes the Gemini-3.1-Pro-Preview detector, and Avg denotes the arithmetic mean of the two detector labels. We report turns $K=1$ to $K=3$, because these are the premise-correct turns in FPCO-Dialog.}
\label{tab:full_corrfp_numbers}
\end{longtable}

\endgroup
\twocolumn

\end{document}